\documentclass[10pt,twocolumn,letterpaper]{article}

\usepackage{cvpr}              %

\usepackage{url}
\usepackage{graphicx}
\usepackage{booktabs} 
\usepackage{multirow}
\usepackage{amsthm}
\usepackage{subcaption}
\usepackage{wrapfig}
\usepackage{algorithm}
\usepackage{algpseudocode}
\usepackage{bbm}

\newtheorem{theorem}{Theorem}

\definecolor{cvprblue}{rgb}{0.21,0.49,0.74}
\usepackage[pagebackref,breaklinks,colorlinks,allcolors=cvprblue]{hyperref}
\def\paperID{ } %
\def\confName{CVPR}
\def\confYear{2026}

\title{Temporal Imbalance of Positive and Negative Supervision \\in Class-Incremental Learning}

\author{
    Jinge Ma$$ \quad Fengqing Zhu$$ \\
    Purdue University, West Lafayette, Indiana, U.S.A.\\
    \texttt{\{ma859, zhu0\}@purdue.edu}
}

\begin{document}
\maketitle
\begin{abstract}

With the widespread adoption of deep learning in visual tasks, Class-Incremental Learning (CIL) has become an important paradigm for handling dynamically evolving data distributions. However, CIL faces the core challenge of \emph{catastrophic forgetting}, often manifested as a prediction bias toward new classes. Existing methods mainly attribute this bias to intra-task class imbalance and focus on corrections at the classifier head. In this paper, we highlight an overlooked factor—\textbf{temporal imbalance}—as a key cause of this bias. Earlier classes receive stronger negative supervision toward the end of training, leading to asymmetric precision and recall. We establish a temporal supervision model, formally define temporal imbalance, and propose \textbf{Temporal-Adjusted Loss} (TAL), which uses a temporal decay kernel to construct a supervision strength vector and dynamically reweight the negative supervision in cross-entropy loss. Theoretical analysis shows that TAL degenerates to standard cross-entropy under balanced conditions and effectively mitigates prediction bias under imbalance. Extensive experiments demonstrate that TAL significantly reduces forgetting and improves performance on multiple CIL benchmarks, underscoring the importance of temporal modeling for stable long-term learning.
\end{abstract}

\section{Introduction}
\vspace{-2mm}
\label{intro}

With the rapid development of deep learning in various visual tasks, models in real-world environments are increasingly exposed to dynamically evolving training data distributions. This paradigm is commonly referred to as Continual Learning (CL)~\cite{kirkpatrick2017overcoming,van2019three,Aljundi_2019_CVPR}. In particular, in the \textbf{Class-Incremental Learning (CIL)} setting, new classes are introduced sequentially as tasks, while the majority of old-class data is no longer accessible in future tasks. Such temporal variation in the training distribution brings the core challenge of CIL: catastrophic forgetting~\cite{mccloskey1989catastrophic, french1999catastrophic}. Specifically, when new tasks introduce a large number of new-class samples, the model’s accuracy on earlier classes drops dramatically. Many studies have indicated that one important cause of this phenomenon is the model’s prediction bias toward new classes~\cite{castro2018end, wu2019large}.  
\begin{figure}[ht!]
    \centering
    \includegraphics[height=4.5cm,keepaspectratio]{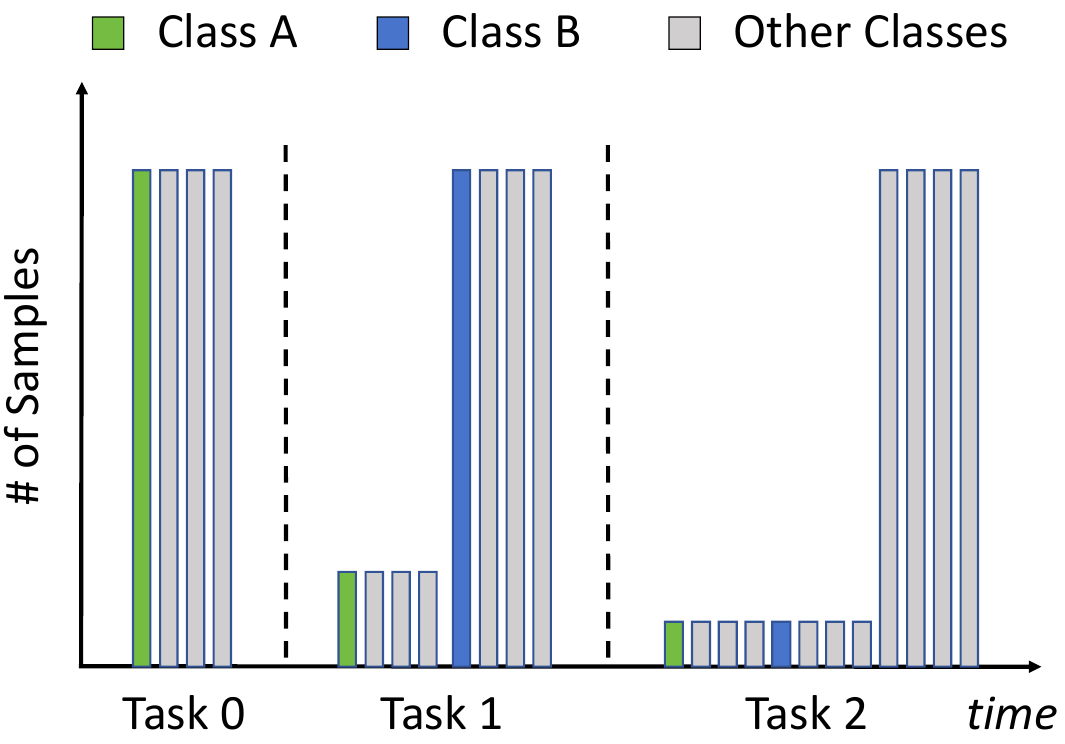}
    \caption{Illustration of temporal imbalance, which differs from class imbalance. 
    In Task~2, although old classes A and B have the same number of training samples, 
    the earlier old class suffers from more severe forgetting than the later one.}
    \label{fig:motivation}
\end{figure}

To alleviate this issue, many methods treat it as a consequence of class imbalance between new and old classes within each task. As a result, numerous CIL approaches borrow ideas from long-tailed learning. In general, existing methods for reducing prediction bias toward new classes can be roughly grouped into three categories. 
\textbf{(1) Balanced fine-tuning}, which retrains the classifier on a small balanced validation set to remove the bias in classifier heads~\cite{wu2019large}. 
\textbf{(2) Prototype-based classifier heads}, which replace the standard fully-connected(FC) head with class prototypes in the feature space and classify samples by their similarity (such as cosine) to these prototypes, thereby reducing the dominance of new-class weights~\cite{rebuffi2017icarl}.
\textbf{(3) Output layer calibration}, which directly adjusts the logits or classifier's parameters through weight alignment or normalization~\cite{ yan2021dynamically}. A more detailed literature review is provided in Sec.~\ref{related}. 

Existing techniques provide effective remedies, however, they mainly operate at the classifier head and explain bias only from the perspective of class imbalance. In this paper, we argue that attributing the prediction bias toward new classes solely to class imbalance between old and new classes is an oversimplification.
For example, as shown in Fig~\ref{fig:motivation}(a), in Task 2, suppose that classes A and B appear with the same frequency, so there is no class imbalance between them. Yet, if most positive samples of class A were provided in Task 0 while those of class B were mainly concentrated in Task 1, then class A will have experienced stronger negative supervision at the beginning of Task 2. This demonstrates that even when classes appear balanced in the current task, differences in the historical distribution of positives still lead to unequal supervision biases between old classes. As a consequence, even though there is no class imbalance among these old classes in the last task, earlier old classes generally suffer stronger negative supervision pressure, often resulting in higher precision but lower recall (see Sec.~\ref{temporal_imbalance} for a detailed analysis).  In addition, the temporal order of training data in CIL introduces a systematic bias, which affects all model parameters, not limited to the classification head.

To address these issues, and to distinguish temporal imbalance from intra-task class imbalance, we model CIL as a problem of temporal imbalance between positive and negative supervision across the entire training process. We propose a time-order-sensitive loss function, Temporal Adjusted Loss (TAL). Specifically, TAL applies a time-decayed memory kernel to convolve the positive and negative supervision sequences of each class, thereby maintaining a vector $Q$ that characterizes the temporal positive supervision strength. Based on $Q$, TAL adaptively reweights the negative supervision in the cross-entropy loss, ensuring that the sensitivity to negative supervision is proportional to the class’s temporal positive supervision strength. In this way, TAL dynamically allocates negative pressure according to each class’s current temporal status.  

In summary, our main contributions are as follows:
\begin{itemize}
    \item We establish a temporal supervision model for CIL, define the problem of temporal imbalance in CIL, and propose Temporal Adjusted Loss (TAL) to mitigate it. 
    \item We provide both theoretical and empirical analyses of TAL, derive the steady-state properties of the vector $Q$, and further introduce a frequency-aligned scaling factor to ensure optimization stability. 
    \item We conduct extensive experiments, demonstrating that TAL effectively reduces forgetting across multiple scenarios.
\end{itemize}

\section{Related Works}
\label{related}
\subsection{Class-Incremental Learning}

Class-Incremental Learning (CIL) is one of the core scenarios of continual learning. Existing methods mainly focus on alleviating catastrophic forgetting: replay-based~\cite{riemer2018learning,chaudhry2019tiny, tiwari2022gcr} approaches maintain an exemplar memory to revisit old classes, regularization-based approaches~\cite{kirkpatrick2017overcoming, li2017learning, lee2019overcoming} constrain parameter updates according to their importance to reduce the damage to old knowledge, and dynamic architecture approaches~\cite{yan2021dynamically,zhou2022model,zheng2025task} expand the network capacity to adapt to new tasks. As discussed in the introduction, to mitigate the prediction bias towards new classes, three common techniques have been proposed as specific strategies within these approaches. \textbf{Balanced fine-tuning} retrains the classifier on a small balanced validation dataset to reduce bias~\cite{wu2019large,douillard2022dytox,wang2022beef}, but it introduces additional training time overhead.
Second, \textbf{Prototype-based classifiers} compute classification scores based on the similarity between a sample and each class prototype, rather than relying on raw logits, thereby alleviating the dominance bias of new-class weights~\cite{rebuffi2017icarl,hou2019learning,douillard2020podnet,zhou2025revisiting}. However, these methods operate only at the classifier head and fail to address the systematic bias caused by representation drift in the backbone.
Third, \textbf{output layer calibration} directly rescales or aligns the logits or classifier weights~\cite{belouadah2019il2m,zhao2020maintaining,yan2021dynamically,wang2022foster, zhou2022model,zheng2025task}. Although lightweight, it is a post-hoc  fix only in the fully-connected layers and could not be applied to cosine-based classifiers.  

Overall, these methods generally attribute prediction bias to the class imbalance between new and old classes and only correct the classifier head. As a result, they cannot alleviate the heterogeneous forgetting degrees across different old classes, nor do they address biases beyond the classifier head. In contrast, our loss function performs temporal modeling of the supervision signals for each class, enabling fine-grained differentiation across classes and systematically correcting the biases that CIL introduces into the overall model.

\subsection{Temporal Modeling in Data Streams}

When dealing with data streams, many fields rely on temporal modeling mechanisms to explain the phenomenon that ``recent data has a greater impact on the present state.'' 
In time series forecasting, exponential smoothing captures trends by assigning decayed weights to historical observations and has been widely applied in economics and financial forecasting~\cite{holt2004forecasting,winters1960forecasting,hyndman2002state,hyndman2018forecasting}. 
For example, the Holt--Winters seasonal exponential smoothing method handles time series with both trend and seasonality.  
In reinforcement learning, eligibility traces and TD($\lambda$)~\cite{sutton1988learning,sutton1998reinforcement,seijen2014true}, as well as their modern extensions such as Retrace($\lambda$)~\cite{munos2016safe} and Expected Eligibility Traces~\cite{van2021expected}, use \textbf{exponential decay} to assign higher ``weight'' to recent state--action pairs, thereby addressing the problem of delayed rewards.  
In online learning and concept drift detection, sliding windows and exponential decay are also widely adopted~\cite{gama2004learning,baena2006early,gama2014survey,gemaque2020overview}. For instance, ADWIN~\cite{bifet2007learning} detects distributional changes in data streams by dynamically adjusting the window size, automatically expanding or shrinking it in response to drift.  
In recommender systems, time-aware collaborative filtering methods incorporate temporal decay or windowing on users' historical behaviors to better model evolving user interests, such as the use of piecewise decay functions for rating prediction~\cite{wu2010time,vinagre2012time}.  

In summary, these approaches share the intuition that recent data should receive larger weights than distant data. However, they are not primarily designed for class-incremental learning. To the best of our knowledge, there has been limited explicit temporal modeling of positive–negative supervision at the loss level in CIL; we take a step in this direction by formalizing temporal imbalance and proposing a remedy for prediction bias.

\section{Preliminaries}
\subsection{CIL Settings}
In CIL, we can denote the training dataset for task $t$ as
\begin{equation}
\mathcal{D}_t = \bigcup_{k\in K_t}\{(x_{t,k,i},\,y_{t,i})\}_{i=1}^{n_{t,k}}
\end{equation}
where $x_{t,k,i}$ and $y_{t,i}$ represent the $i$-th image and its corresponding label for class $k$ in task $t$, respectively. The value $n_{t,k}$ denotes the number of samples for class $k$ in task $t$, and $K_t$ is the set of all class indices present in the training data of task $t$ (including replay exemplars if any).  

Assume the model parameters are $\theta$ and the model has learned the first $T$ tasks and is now learning task $T+1$. Since the datasets \( \bigcup_{t=1}^T \mathcal{D}_t \) for all previous tasks are no longer available, the model can only adapt its existing parameters to fit
\(\mathcal{D}_{T+1}\), which typically leads to \textbf{catastrophic forgetting}—the model's performance on new tasks improves while its accuracy on earlier tasks deteriorates significantly.

\subsection{Positive and Negative Supervision}
For a single sample $(x,y)$, let $z\in\mathbb{R}^{C}$ denote the pre-softmax output logits of the classifier, with $z_k$ the $k$-th logit and $C$ the total number of classes. The cross-entropy (CE) loss can be written as

\begin{equation}
\begin{aligned}  
\ell_{\text{CE}}(y,z)
&= -\log \frac{e^{z_y}}{e^{z_y}+\sum_{k\ne y} e^{z_k}} \\
&= \underbrace{\log\!\Big(e^{z_y}+\sum_{k\ne y}e^{z_k}\Big)}_{\text{negative supervision}}
\;-\;
\underbrace{z_{y}}_{\text{positive supervision}}
\label{eq:ce_decomp}
\end{aligned}
\end{equation}

Thus, to minimize the $\ell_{\text{CE}}$ for $x$, as shown in Eq.~\ref{eq:ce_decomp}, the ``positive supervision'' pushes up the true-class logit, while the ``negative supervision'' corresponds to the softmax denominator in which all classes exert suppressive pressure. In this sense, for a given class, its positive samples are those belonging to the class (providing positive supervision), while its negative samples are those not belonging to it (providing negative supervision).

\section{Methodology}
\label{method}
This section introduces how we interpret and rectify the model prediction bias in CIL from the perspectives of temporal order and supervision polarity. 
We introduce a temporal modeling mechanism to track, for each class $k$, its \textbf{temporal positive supervision strength} $Q_k$, 
which characterizes the extent to which the class has been reinforced by positive samples in recent steps. 
$Q_k$ is recursively updated through an exponentially decaying memory kernel function $f[n]$, 
reflecting the fact that more recent samples have a stronger influence on the current supervision polarity of that class. 
Based on $Q_k$, we propose the \textbf{Temporal-Adjusted Loss (TAL)}, 
which dynamically adjusts the sensitivity to negative supervision according to each class’s current $Q_k$: 
for old classes that have recently lacked positive supervision (with smaller $Q_k$), 
their negative supervision weights are automatically attenuated; 
whereas for new classes that have been sufficiently reinforced (with larger $Q_k$), 
the sensitivity to negative supervision is preserved. 
In addition, we introduce a \textbf{frequency alignment coefficient} $\alpha$, 
which ensures that, for classes with balanced temporal and categorical distributions, 
TAL degenerates into the standard cross-entropy loss.

\subsection{Temporal Modeling of Supervision}

\paragraph{Temporal Positive Supervision Strength $Q$.}
We consider a simple case where one sample arrives per step (i.e., unit batch size). For a given class $k$, at step $N$, we denote its supervision polarity sequence $\{a_k[n]\}_{n=0}^{N-1}$ by:
\begin{equation}
a_k[n]=\begin{cases}
+1, & \text{if the sample at step $n$ is positive to class $k$},\\
-1, & \text{if the sample at step $n$ is negative to class $k$}.
\end{cases}
\label{eq:sup_seq}
\end{equation}

To model temporal effects, we introduce a general decay memory kernel function $f[\cdot]$, where $f[n]$ ($n\in\mathbb N_0$) represents the residual influence on the current step from a sample which was seen $n$ steps ago (equivalently, the influence of a sample after $n$ steps). 

Then, at step $N$, the temporal positive supervision strength vector $Q[N]$ is defined as
\begin{equation}
\begin{aligned}  
&Q[N] := \big(Q_1[N], Q_2[N], \dots, Q_C[N]\big),\\
&Q_k[N] = \sum_{n=0}^{N-1} f[N-1-n]\;a_k[n] \;=\; (f * a_k)[N-1].
\label{eq:Q-def}
\end{aligned}
\end{equation}
where $(f * a_k)$ denotes the discrete convolution between $f[\cdot]$ and the supervision sequence $a_k[\cdot]$.  

Intuitively, our defined $Q_k[N]$ measures the effective positive supervision of class $k$ at step $N$: positive samples increase $Q_k$, while negative samples decrease it. Naturally, $f[n]$ is defined to satisfy $f[n]\ge 0$ and $f[n+1]\le f[n]$ for any $n$, reflecting the fact that more recent training samples have stronger impacts on $Q_k[N]$ due to model's forgetting. The specific choice of $f[\cdot]$ will be discussed in Sec.~\ref{memory_kernel}. A larger $Q_k$ indicates stronger positive supervision to class $k$ at the current step, whereas a smaller $Q_k$ implies that positive supervision for the class is relatively lacking.

\subsection{Temporal Imbalance}
\label{temporal_imbalance}
Consider two classes $A$ and $B$ observed over $N$ steps with the \emph{same} total number of positive samples. Up to step $n$, denote the \textbf{number of cumulative positive samples} of class $k$ as $S_k[n]$. 

\begin{theorem}[Temporal imbalance under equal sample counts]\label{thm:backloading}
Consider two classes $A$ and $B$ with the same total number of positive samples. 
Since $f[n]$ is monotonically decreasing, it follows that
\begin{equation}
\begin{aligned}
&S_A[n]\;\ge\;S_B[n]\quad \forall\,n=0,1,\dots,N-1\\
&\Longrightarrow\quad Q_A[N]\;\le\;Q_B[N].
\label{eq:backorder-ineq}
\end{aligned}
\end{equation}
with strict inequality if $\exists\,n$ such that $S_A[n]>S_B[n]$ and $f[n]$ is strictly decreasing.
\end{theorem}
The proof of Thm.~\ref{thm:backloading} is deferred to the Appendix.

\begin{figure}[ht!]
    \centering

    \begin{subfigure}{0.485\linewidth}
        \centering
        \includegraphics[height=4.05cm,keepaspectratio]{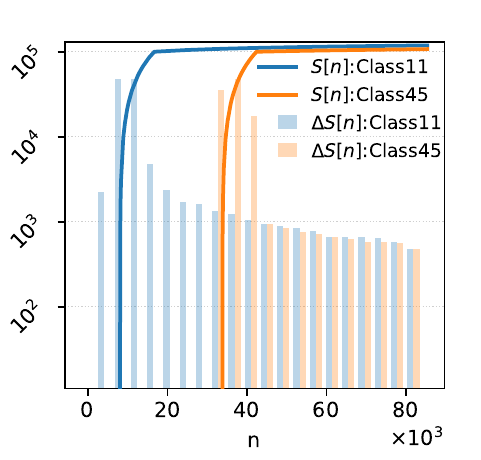}
        \caption{$S[n]$}
    \end{subfigure}
    \hfill
    \begin{subfigure}{0.485\linewidth}
        \centering
        \includegraphics[height=3.8cm,keepaspectratio]{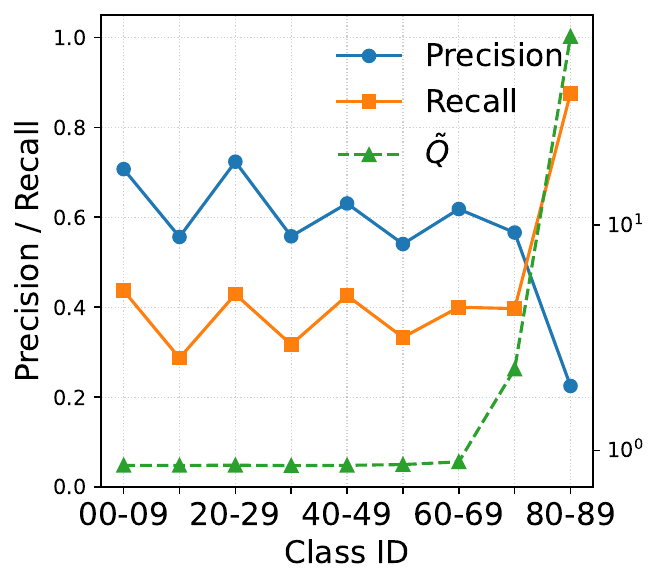}
        \caption{$Q$ vs Precision\&Recall}
    \end{subfigure}

    \par\medskip %

    \begin{subfigure}{0.99\linewidth}
        \centering
        \includegraphics[height=4.05cm,keepaspectratio]{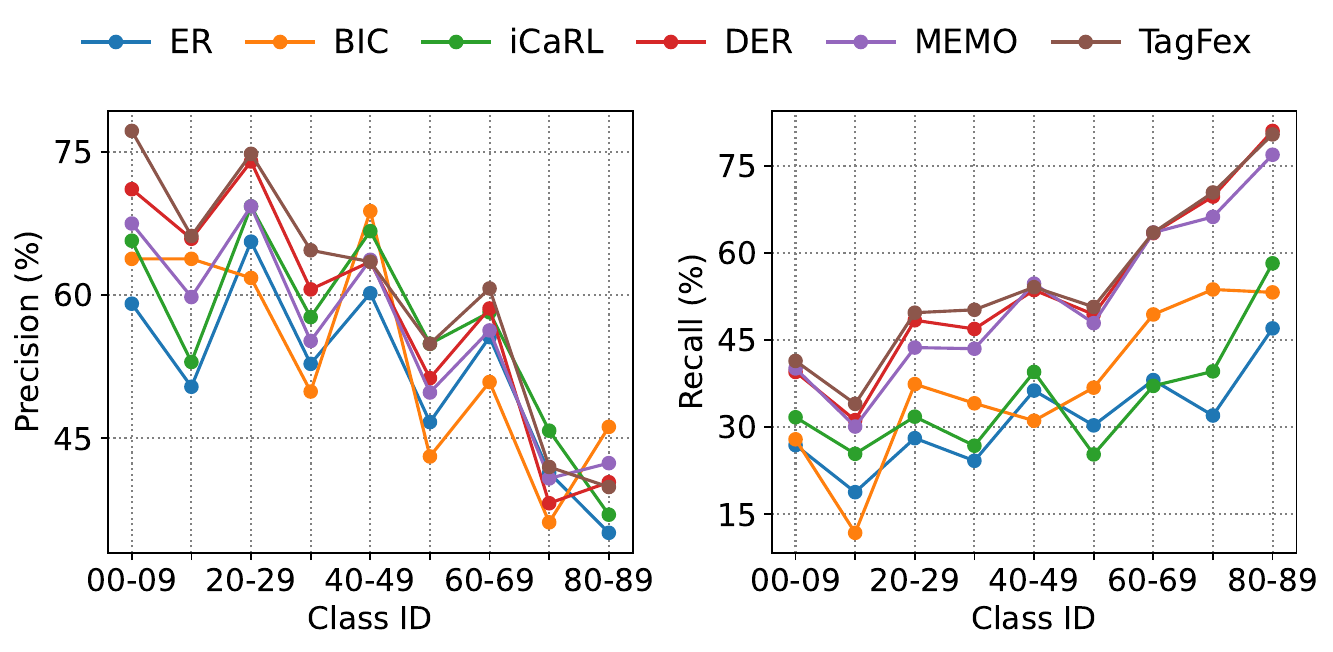}
        \caption{Precision/Recall asymmetry in various baselines}
    \end{subfigure}

    \caption{
    \textbf{(a) An example of Temporal Imbalance:} 
    it illustrates, in the CIL setting, how the cumulative number of positive samples $S[n]$ for two classes evolves as the number of training batches $n$ increases. 
    Since the positive samples of class~11 appear earlier than those of class~45, the result shows that $S_{11}[n] \ge S_{45}[n]$.
    \textbf{(b)\&(c) Effects of Temporal Imbalance:} 
    \textbf{(b)}~Shows, at the end of training, the correlation between the Precision/Recall of each class and its $Q$ value when ER is trained with the cross-entropy loss. 
    For clearer visualization, the normalized value $\tilde{Q} = Q - Q_{\min}$ is used, where $Q_{\min}$ denotes the minimum $Q$ across all classes.  
    \textbf{(c)}~Demonstrates that, at the end of training, the Precision–Recall asymmetry induced by temporal imbalance is a common phenomenon across different CIL methods. 
    Note that all classes shown in the figure are old classes, so there is no class imbalance in the final task.}
    \label{fig:functions}
    \vspace{-4mm}
\end{figure}

\paragraph{Interpretation.}
Thm.~\ref{thm:backloading} formalizes temporal imbalance that is independent of class imbalance: with the same number of positive samples $M$, the class whose positives occur later in time will attain a larger $Q_k[N]$ at the end of training. In the CIL setting, an early class $A$ attains a smaller $Q_A[N]$ than a late class $B$, as shown in Fig.~\ref{fig:functions}(a), i.e., $Q_A[N]<Q_B[N]$. Intuitively, a smaller $Q_k[N]$ means that negative supervision dominates at step $N$, which suppresses the logits of class $k$ and makes predictions for class $A$ concentrate on high-confidence samples, yielding higher precision but lower recall. Conversely, a larger $Q_B[N]$ indicates stronger positive reinforcement, admitting more less confident cases, which improves recall at the cost of precision. As shown in Fig.~\ref{fig:functions}(b), the intuitive effect of temporal imbalance is the asymmetry between precision and recall: at the end of training, earlier classes (e.g., class~A) have smaller $Q$ values, leading to higher precision but lower recall, whereas later classes (e.g., class~B) have larger $Q$ values, resulting in lower precision but higher recall. As illustrated in Fig.~\ref{fig:functions}(c), our empirical studies demonstrate that this precision–recall asymmetry induced by temporal imbalance is a pervasive phenomenon across various existing CIL methods.

\subsection{Choice of the Memory Kernel}
\label{memory_kernel}
In this paper, we follow the exponential temporal decay mechanism in prior work~\cite{seijen2014true}, and adopt the exponential decay memory kernel
\begin{equation}
f[n] = \lambda^{\,n+1}, \qquad 0 < \lambda < 1.
\label{eq:kernel}
\end{equation}
where we refer to $\lambda$ as the \textbf{memory parameter}, which controls how fast the contribution of positive or negative supervision from past samples to $Q_k[N]$, defined in Eq.~\ref{eq:Q-def}, decays over time. A larger $\lambda$ means we suppose the $Q_k[N]$ retains stronger memory of earlier supervision's polarity, whereas a smaller $\lambda$ implies weaker memory. 

Once the hyperparameter $\lambda$ is fixed, we can derive an asymptotic upper bound $Q_{\max}$ for $Q_k[N]$. Consider the case when $N\to\infty$ and all training samples belong to class $k$ (i.e., $a_k[n]\equiv 1$). By the definition of $Q_k[N]$ in Eq.~\ref{eq:Q-def}, we always have $Q_k[N]<Q_{\max}$ and
\begin{equation}
\begin{aligned}
Q_{\max} &= \lim_{N\to\infty} Q_k[N] 
= \lim_{N\to\infty} \sum_{n=0}^{N-1} \lambda^{\,N-n}a_k[n]\\
&= \sum_{m=1}^{\infty} \lambda^{\,m}
= \frac{\lambda}{1-\lambda}.
\label{eq:qmax}
\end{aligned}
\end{equation}

Once the exponential decay memory kernel is fixed, the computation of $Q$ can be viewed as a Markov chain: $Q_k(N\!+\!1)$ depends only on $Q_k[N]$ and the current supervision $a_k[N]$, without explicitly storing the entire sequence $\{a_k[n]\mid n=0,\dots,N-1\}$. Specifically,
\begin{equation}
\begin{aligned}
Q_k(N+1) 
&= \sum_{n=0}^{N} \lambda^{\,N+1-n}a_k[n] \\
&= \lambda \Big(\sum_{n=0}^{N-1} \lambda^{\,N-n}a_k[n]\Big) + \lambda a_k[N] \\
&= \lambda\big(Q_k[N]+a_k[N]\big).
\label{eq:markov}
\end{aligned}
\end{equation}
This recursive form reduces the calculation of $Q_k[N]$ from $\mathcal{O}[N]$ to $\mathcal{O}(1)$ time and memory complexity, since only $Q_k[N]$ and $a_k[N]$ are required.

\subsection{Our Method: Temporal-Adjusted Loss}
\label{TAL}

From the above analysis, we observe that temporal imbalance causes old classes to receive excessive negative supervision at the end of training, leading to model's bias toward new classes. To address this issue, we design \textbf{Temporal-Adjusted Loss (TAL)}. Our motivation is to dynamically adjust the sensitivity to negative supervision from the perspective of the loss function, based on each class's current $Q_k[N]$ value. In this way, classes with insufficient positive supervision are protected from excessive negative pressure, while classes with sufficient positive supervision remain highly sensitive to negative supervision.

\paragraph{TAL Function.}
For a single training sample $(x,y)$ with logits $z\in\mathbb{R}^C$, given the current temporal positive supervision strength $Q[N]$, we define its TAL loss as:
\begin{equation}
\begin{aligned}
&\ell_{\text{TAL}}(y,z,Q[N])
= -\log\!\left(
\frac{e^{z_{y}}}{e^{z_{y}}+\alpha \sum_{k\ne y}w\!\big(Q_k[N]\big)\,e^{z_{k}}}
\right), \\
&\text{where } w\!\big(Q_k[N]\big)=\left(\frac{Q_k[N]}{Q_{\max}}\right)^{r}
\label{eq:tal_loss}
\end{aligned}
\end{equation}

By comparing Eq.~\ref{eq:ce_decomp} with Eq.~\ref{eq:tal_loss}, we observe that besides the true label $y$ and the model prediction $z$, TAL additionally incorporates the current $Q[N]$, thereby introducing temporal information. Specifically, the true class $e^{z_y}$ directly participates in the loss calculation, which is identical to CE loss. For non-true classes, the $e^{z_k}$ ($k \neq y$) is multiplied by a weight function $w(\cdot)$ positively correlated with $Q_k[N]$, and then further scaled by a frequency alignment parameter $\alpha$ before being included in the loss. The exponent $r$($r>0$) in $w(\cdot)$ controls the steepness of this weighting: larger $r$ makes $w(Q_k[N])$ grow more sharply with $Q_k[N]$. For a class $k$ ($k \neq y$), the closer $Q_k[N]$ is to the upper bound $Q_{\max}$, the larger $w(Q_k[N])$ becomes, making it more sensitive to negative supervision. Conversely, the closer $Q_k[N]$ is to $0$, the less sensitive it is to negative supervision. The range of $Q_k[N]$, the range of $w(\cdot)$, and the derivation of $\alpha$ will be discussed later.

\vspace{-3mm}
\paragraph{Update of $Q[N]$.}  
With the introduction of the weight function $w(\cdot)$, the recursive update of $Q_k$ in Eq.~\ref{eq:markov} is modified: the effect of negative supervision is attenuated, while the strength of positive supervision remains unchanged. The update rule is
\begin{equation}
Q_k[N+1] = \lambda\Big(Q_k[N] + 
\begin{cases}
a_k[N], & a_k[N] = +1, \\[4pt]
w\!\big(Q_k[N]\big)\,a_k[N], & a_k[N] = -1,
\end{cases}
\Big).
\label{eq:markov-tal}
\end{equation}
This ensures that the update of $Q$ is consistent with the attenuation of negative supervision in TAL.

\subsection{Parameter Analysis}
\label{sec:parameter-analysis}

\paragraph{Boundary conditions of $Q$ and $w(\cdot)$.}
From Eq.~\ref{eq:qmax} and ~\ref{eq:markov-tal}, the temporal positive supervision strength satisfies
\begin{equation}
0 \;\le\; Q_k[N] \;<\; Q_{\max}=\frac{\lambda}{1-\lambda}.
\label{eq:Q_range}
\end{equation}
The lower bound is attainable (e.g., when class $k$ receives no positive supervision), while the upper bound is not attained for $0<\lambda<1$. Since $Q_k[N]\in[0,Q_{\max})$, according to the definition of $w(Q_k[N])$ in Eq.~\ref{eq:tal_loss}, we have $0 \;\le\; w(Q_k[N]) \;<\; 1$. At the lower bound $Q_k[N]=0$, we have $w=0$, meaning class $k$ is completely insensitive to negative supervision because of the lack of recent positive samples, effectively screened out in the loss computation. At the upper limit $Q_k[N]\to Q_{\max}$, we have $w\to 1$, recovering full sensitivity to negative supervision. Formal proofs based on the recursion in Eq.~\ref{eq:markov-tal} are deferred to Appendix.

\paragraph{Frequency alignment parameter $\alpha$.}
The role of $\alpha$ is to ensure that, on a temporally uniform (e.g., randomly shuffled) and class-balanced dataset with $C$ classes (each class has prior $p=1/C$), TAL degenerates to the vanilla cross-entropy loss. In other words, in the absence of temporal imbalance, the negative supervision weight coefficient for each class $k$ on their negative samples should revert to $1$. Sepecifically, we require $\alpha \cdot w(Q^\ast) = 1$, where $Q^\ast$ denotes the steady-state temporal positive supervision strength under balanced conditions.  

Taking expectation in Eq.~\ref{eq:markov-tal} and imposing the steady-state condition $Q_k[N+1]=Q_k[N]=Q^\ast$ yields the calibration equation
\begin{equation}
\Big(1 - \frac{1}{C}\Big)(x^\ast)^r + x^\ast - \frac{1}{C} = 0,
\qquad 
\alpha = \frac{1}{(x^\ast)^r},
\label{eq:steady-alpha}
\end{equation}

where $x^\ast = Q^\ast/Q_{\max}$. This equation admits a unique solution $x^\ast \in (0,1)$ (\textbf{proof in Appendix}). Notably, $\alpha$ is solely determined by the number of classes $C$ and the exponent $r$.

\paragraph{Summary of parameters.}
In summary, the temporal modeling introduces several parameters: 
$\lambda$ is the memory parameter in Eq.~\ref{eq:kernel}, modeling the decay speed of past supervision; 
$Q_k[N]$ in Eq.~\ref{eq:Q-def} tracks the temporal positive supervision strength for class $k$; 
$w(Q_k[N])$ in Eq.~\ref{eq:tal_loss} rescales the negative supervision to class $k$ based on $Q_k[N]$, with its steepness controlled by exponent $r$; 
and $\alpha$ in Eq.~\ref{eq:steady-alpha} is the frequency alignment parameter that calibrates the overall balance between positive and negative supervision under temporally uniform and class-balanced conditions. 
Among these, \textbf{only $\lambda$ and $r$ are hyperparameters} to be chosen by the practitioner, while $Q_k[N]$, $w(Q_k[N])$, and $\alpha$ are uniquely determined once $\lambda$ and $r$ are specified.

\begin{table*}[!ht]
\centering
\caption{Average accuracy ($A_{Mean}$) and last accuracy ($A_{Last}$) under 10-task and 20-task settings across three datasets. “+ Ours” indicates the baseline method with the Temporal-Adjusted Loss applied. \textbf{All methods use a replay buffer of 20 exemplars per class.}}
\resizebox{1\linewidth}{!}{
\begin{tabular}{l cc cc cc cc cc}
\toprule
\multirow{2}{*}{Method} 
  & \multicolumn{4}{c}{CIFAR-100} 
  & \multicolumn{4}{c}{ImageNet-100} 
  & \multicolumn{2}{c}{Food101} \\ 
\cmidrule(lr){2-5} \cmidrule(lr){6-9} \cmidrule(lr){10-11}
  & \multicolumn{2}{c}{10-task} 
  & \multicolumn{2}{c}{20-task} 
  & \multicolumn{2}{c}{10-task} 
  & \multicolumn{2}{c}{20-task} 
  & \multicolumn{2}{c}{10-task} \\
\cmidrule(lr){2-3}\cmidrule(lr){4-5}
\cmidrule(lr){6-7}\cmidrule(lr){8-9}
\cmidrule(lr){10-11}
  & $A_{Mean}$ & $A_{Last}$
  & $A_{Mean}$ & $A_{Last}$
  & $A_{Mean}$ & $A_{Last}$
  & $A_{Mean}$ & $A_{Last}$
  & $A_{Mean}$ & $A_{Last}$ \\ 
\midrule
iCaRL~\cite{rebuffi2017icarl}       & 58.76{\scriptsize$\pm$0.12} & 45.39{\scriptsize$\pm$0.22} & 55.98{\scriptsize$\pm$0.14} & 43.20{\scriptsize$\pm$0.24} & 43.71{\scriptsize$\pm$0.12} & 24.38{\scriptsize$\pm$0.24} & 33.45{\scriptsize$\pm$0.14} & 17.62{\scriptsize$\pm$0.28} & 63.46{\scriptsize$\pm$0.12} & 50.13{\scriptsize$\pm$0.22} \\
+ Ours       & \textbf{60.82}{\scriptsize$\pm$0.10} & \textbf{47.36}{\scriptsize$\pm$0.18} & \textbf{58.68}{\scriptsize$\pm$0.12} & \textbf{44.79}{\scriptsize$\pm$0.20} & \textbf{52.19}{\scriptsize$\pm$0.10} & \textbf{32.78}{\scriptsize$\pm$0.20} & \textbf{42.54}{\scriptsize$\pm$0.12} & \textbf{23.90}{\scriptsize$\pm$0.24} & \textbf{68.57}{\scriptsize$\pm$0.10} & \textbf{56.03}{\scriptsize$\pm$0.18} \\
\midrule
FOSTER~\cite{wang2022foster}       & 62.04{\scriptsize$\pm$0.42} & 49.30{\scriptsize$\pm$0.64} & 56.99{\scriptsize$\pm$0.46} & 47.34{\scriptsize$\pm$0.66} & 48.39{\scriptsize$\pm$0.46} & 20.88{\scriptsize$\pm$0.70} & 35.67{\scriptsize$\pm$0.48} & 15.28{\scriptsize$\pm$0.72} & 63.88{\scriptsize$\pm$0.46} & 52.60{\scriptsize$\pm$0.66} \\
+ Ours       & \textbf{64.80}{\scriptsize$\pm$0.40} & \textbf{52.76}{\scriptsize$\pm$0.60} & \textbf{58.21}{\scriptsize$\pm$0.42} & \textbf{49.28}{\scriptsize$\pm$0.64} & \textbf{51.49}{\scriptsize$\pm$0.42} & \textbf{23.38}{\scriptsize$\pm$0.66} & \textbf{38.07}{\scriptsize$\pm$0.46} & \textbf{19.72}{\scriptsize$\pm$0.70} & \textbf{64.61}{\scriptsize$\pm$0.42} & \textbf{53.90}{\scriptsize$\pm$0.60} \\
\midrule
MEMO~\cite{zhou2022model}       & 61.68{\scriptsize$\pm$0.16} & 51.70{\scriptsize$\pm$0.30} & 56.98{\scriptsize$\pm$0.18} & 48.93{\scriptsize$\pm$0.34} & 47.01{\scriptsize$\pm$0.16} & 34.98{\scriptsize$\pm$0.34} & 34.73{\scriptsize$\pm$0.18} & 22.78{\scriptsize$\pm$0.36} & 62.25{\scriptsize$\pm$0.16} & 54.75{\scriptsize$\pm$0.30} \\
+ Ours       & \textbf{63.81}{\scriptsize$\pm$0.12} & \textbf{52.85}{\scriptsize$\pm$0.28} & \textbf{58.04}{\scriptsize$\pm$0.16} & \textbf{49.86}{\scriptsize$\pm$0.30} & \textbf{51.05}{\scriptsize$\pm$0.12} & \textbf{37.28}{\scriptsize$\pm$0.30} & \textbf{38.76}{\scriptsize$\pm$0.16} & \textbf{26.16}{\scriptsize$\pm$0.34} & \textbf{64.73}{\scriptsize$\pm$0.12} & \textbf{55.83}{\scriptsize$\pm$0.28} \\
\midrule
DER~\cite{yan2021dynamically}       & 63.53{\scriptsize$\pm$0.19} & 50.75{\scriptsize$\pm$0.35} & 57.47{\scriptsize$\pm$0.17} & 50.34{\scriptsize$\pm$0.31} & 52.25{\scriptsize$\pm$0.25} & 40.28{\scriptsize$\pm$0.49} & 43.95{\scriptsize$\pm$0.19} & 25.82{\scriptsize$\pm$0.35} & 69.20{\scriptsize$\pm$0.23} & 61.42{\scriptsize$\pm$0.47} \\
+ Ours       & \textbf{66.33}{\scriptsize$\pm$0.17} & \textbf{53.82}{\scriptsize$\pm$0.31} & \textbf{60.34}{\scriptsize$\pm$0.13} & \textbf{53.49}{\scriptsize$\pm$0.29} & \textbf{54.57}{\scriptsize$\pm$0.23} & \textbf{42.62}{\scriptsize$\pm$0.43} & \textbf{47.80}{\scriptsize$\pm$0.17} & \textbf{29.79}{\scriptsize$\pm$0.31} & \textbf{71.58}{\scriptsize$\pm$0.19} & \textbf{63.98}{\scriptsize$\pm$0.41} \\
\midrule
TagFex~\cite{zheng2025task}       & 65.97{\scriptsize$\pm$0.20} & 55.99{\scriptsize$\pm$0.41} & 59.55{\scriptsize$\pm$0.20} & 51.48{\scriptsize$\pm$0.41} & 54.73{\scriptsize$\pm$0.35} & 41.70{\scriptsize$\pm$0.59} & 47.32{\scriptsize$\pm$0.23} & 31.36{\scriptsize$\pm$0.35} & 71.55{\scriptsize$\pm$0.31} & 66.67{\scriptsize$\pm$0.53} \\
+ Ours       & \underline{\textbf{68.68}}{\scriptsize$\pm$0.18} & \underline{\textbf{57.91}}{\scriptsize$\pm$0.37} & \underline{\textbf{62.34}}{\scriptsize$\pm$0.18} & \underline{\textbf{54.70}}{\scriptsize$\pm$0.37} & \underline{\textbf{57.05}}{\scriptsize$\pm$0.31} & \underline{\textbf{43.01}}{\scriptsize$\pm$0.55} & \underline{\textbf{51.08}}{\scriptsize$\pm$0.19} & \underline{\textbf{35.78}}{\scriptsize$\pm$0.31} & \underline{\textbf{74.57}}{\scriptsize$\pm$0.29} & \underline{\textbf{68.64}}{\scriptsize$\pm$0.49} \\
\bottomrule
\end{tabular}
}
\label{tab:replay_all}
\end{table*}

\section{Experiments}

In this section, we conduct two groups of experiments. 
In Sec.~\ref{main_exp}, we integrate TAL into both classical and most recent CIL methods on three benchmark datasets to observe the improvements brought by TAL. In Sec.~\ref{ablation}, we perform ablation studies on CIFAR-100 to investigate the effects of hyperparameters $\lambda$, $r$. For all methods, the rehearsal samples are selected using the herding strategy~\cite{rebuffi2017icarl}.

\subsection{Experimental setup}
\textbf{Datasets.}
We conduct experiments on CIFAR-100~\cite{krizhevsky2009learning}, ImageNet-100~\cite{russakovsky2015imagenet}, and Food101~\cite{bossard2014food}. 
CIFAR-100 consists of 50,000 training images and 10,000 test images across 100 classes, with each image being a small $32\times32$ color image. 
ImageNet-100 is a 100-class subset of ImageNet, containing 1,300 training samples and 50 test samples per class. Food101 includes 101 food categories with 75,750 training and 25,250 test images.

\noindent\textbf{Implementation Details.}
We conduct experiments using a public class incremental learning toolboxes PyCIL~\cite{zhou2024continual}. To ensure reproducibility, We provide performance results with standard deviations computed over 5 runs. For model backbones, we use ResNet32 on CIFAR-100, and ResNet18 on ImageNet-100 and Food101.  All other hyperparameters for the CIL methods and the optimizer are left at the official codebase’s default settings.

\subsection{Results and Visualizations}
\label{main_exp}
\begin{figure}[h] %
  \vspace{-2mm}
  \centering
  \includegraphics[width=\linewidth]{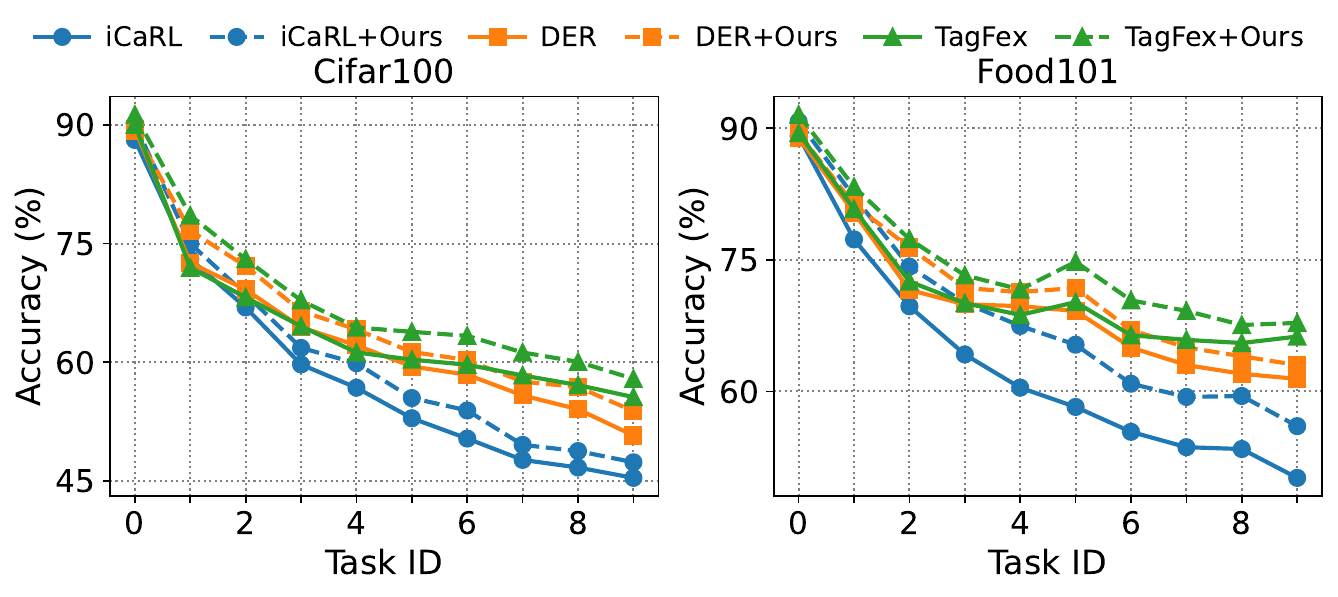}
  \caption{Forgetting curves with representative CIL baselines and datasets.}

  \label{fig:forgetting_curves}
\end{figure}

We adopt iCaRL~\cite{rebuffi2017icarl}, FOSTER~\cite{wang2022foster}, DER~\cite{yan2021dynamically}, MEMO~\cite{zhou2022model}, and TagFex~\cite{zheng2025task} as baseline methods. In Sec.~\ref{main_exp}, the replay buffer size is\textbf{ fixed at 20 samples per class across all methods}. For iCaRL (with or without TAL), we follow its original paper and report the \textbf{NME (Nearest Mean of Exemplars) accuracy}, while for the other methods, we directly compute accuracy based on the output logits. As shown in Tab.~\ref{tab:replay_all} and Fig.~\ref{fig:forgetting_curves}, TAL consistently improves the performance of various baselines across different task configurations, regardless of their complexity. Notably, after applying TAL, even the most basic baseline, iCaRL, surpasses more advanced methods such as FOSTER and MEMO in several setups, including CIFAR-100 (20-task), ImageNet-100 (10-task), and Food101 (10-task).

\begin{figure}[h]
    \centering
    \includegraphics[width=0.48\textwidth, height=0.33\textheight]{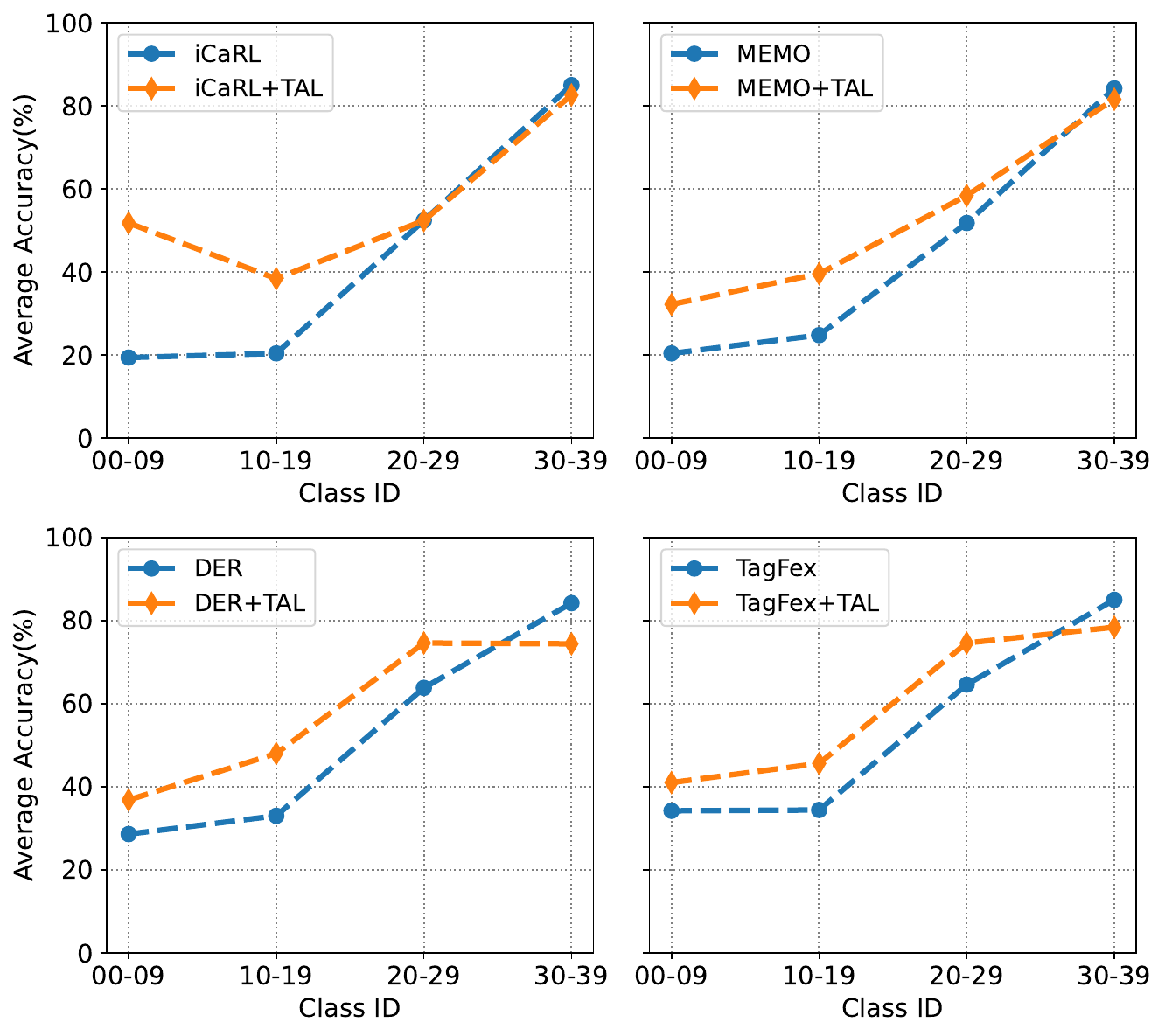}
    \caption{This figure shows the performance of different baselines on each class when Task id = 3 under the 10-task setting on ImageNet-100. It can be observed that earlier classes exhibit lower average class accuracy (i.e., recall), similar to Fig~\ref{fig:functions}(c). TAL reduces the sensitivity of old classes to negative supervision while increasing that of new classes. As a result, TAL improves the recall of most old classes at the cost of lowering the recall of new classes slightly.}
    \label{fig:baselines}
    \vspace{-5mm}
\end{figure}

To better illustrate the reasons behind the performance improvements brought by TAL, we visualize the results of four baseline methods (iCaRL, DER, MEMO, and TagFex) on ImageNet-100 under a fixed learning order. Fig.~\ref{fig:baselines} shows the class accuracy curves over class IDs, arranged according to the learning sequence, after the completion of the final task. It can be observed that even with weight alignment, the recall (i.e., class accuracy) of earlier classes remains significantly lower due to the lack of temporal modeling, while relatively newer classes achieve much higher recall. After introducing TAL, however, the model no longer applies the same level of protection to earlier and more recent old classes with equal numbers of exemplars. Instead, it adjusts its prediction bias more reasonably according to the temporal order of classes.

\begin{figure}[h] %
  \vspace{-2mm}
  \centering
  \includegraphics[width=\linewidth]{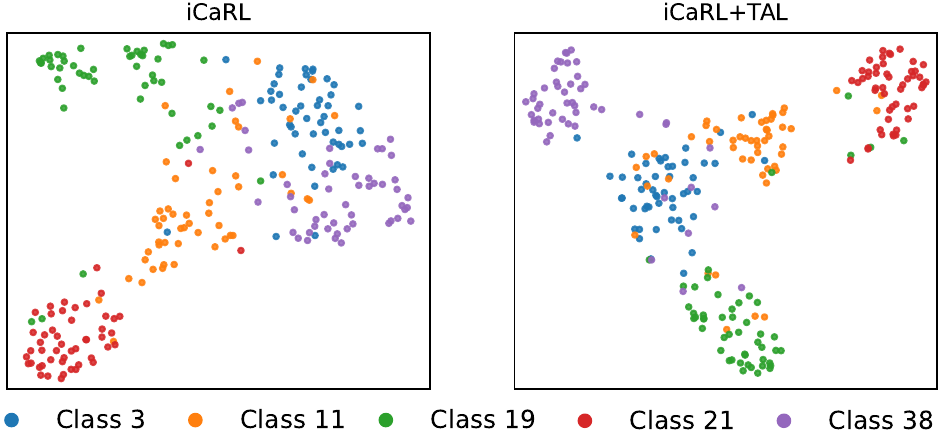}
  \caption{This figure illustrates the impact of applying TAL on the feature space of iCaRL. We visualize the distribution of test images from several new and old classes in the ResNet18 feature space on ImageNet100 at Task ID = 3.}
  \label{fig:feature_space}
  \vspace{-3mm}
\end{figure}

In Fig.~\ref{fig:feature_space}, we illustrate the impact of TAL on the feature space of the backbone network. We use UMAP~\cite{mcinnes2018umap} to visualize the spatial distribution of feature vectors from five classes sampled across different tasks of iCaRL. It can be observed that earlier classes(such as Class 3 and Class 11) are more prone to being mixed with other classes in the feature space, and their regions are gradually occupied by newer classes(such as Class 38). After introducing TAL, this phenomenon is mitigated. This indicates that TAL not only provides corrections at the classifier level but also has effects that are not limited to linear classification heads.

\subsection{Ablation Studies}

\label{ablation}
We investigate the effect of TAL’s two hyperparameters—the memory parameter $\lambda$ and the steepness exponent $r$—on model performance, using \textbf{Experience Replay (ER)} as the baseline. All experiments are conducted on the CIFAR100\&10-task benchmark with a ResNet32 backbone under different replay buffer sizes.

\begin{table}[!ht]
\centering
\caption{Ablations on $\lambda$ and $r$. We report both average accuracy ($A_{\text{Mean}}$) and last accuracy ($A_{\text{Last}}$).}
\resizebox{1\linewidth}{!}{
\begin{tabular}{c|cccccccc}
\toprule
\multirow{3}{*}{$r$} 
 & \multicolumn{8}{c}{$\lambda$} \\
\cmidrule(lr){2-9}
 & \multicolumn{2}{c}{0.99} 
 & \multicolumn{2}{c}{0.995} 
 & \multicolumn{2}{c}{0.999} 
 & \multicolumn{2}{c}{0.9995} \\
\cmidrule(lr){2-3} \cmidrule(lr){4-5} \cmidrule(lr){6-7} \cmidrule(lr){8-9}
 & $A_{\text{Mean}}$ & $A_{\text{Last}}$
 & $A_{\text{Mean}}$ & $A_{\text{Last}}$
 & $A_{\text{Mean}}$ & $A_{\text{Last}}$
 & $A_{\text{Mean}}$ & $A_{\text{Last}}$ \\
\midrule
0.2 & 61.30 & 44.39 & 61.06 & 44.05 & 60.69 & 43.54 & 60.26 & 43.18 \\
0.5 & 62.12 & 45.36 & 61.27 & 45.05 & 61.72 & 44.50 & 61.36 & 44.28 \\
1.0 & \textbf{62.60} & \textbf{46.13} & \underline{\textbf{63.36}} & \underline{\textbf{47.64}} & 62.46 & \textbf{46.66} & 61.79 & 45.44 \\
2.0 & 62.51 & 45.38 & 60.24 & 45.80 & \textbf{62.85} & 46.16 & \textbf{62.50} & \textbf{44.85} \\
5.0 & 56.70 & 36.90 & 57.15 & 37.97 & 58.62 & 38.83 & 58.08 & 38.94 \\
\midrule
CE & 59.96 & 42.39 & - & - & - & - & - & - \\
\bottomrule
\end{tabular}
}
\label{tab:hyper_lambda_r}
\vspace{-2mm}
\end{table}

\begin{figure}[h]
    \centering
    \includegraphics[width=1\linewidth, height=0.5\linewidth]{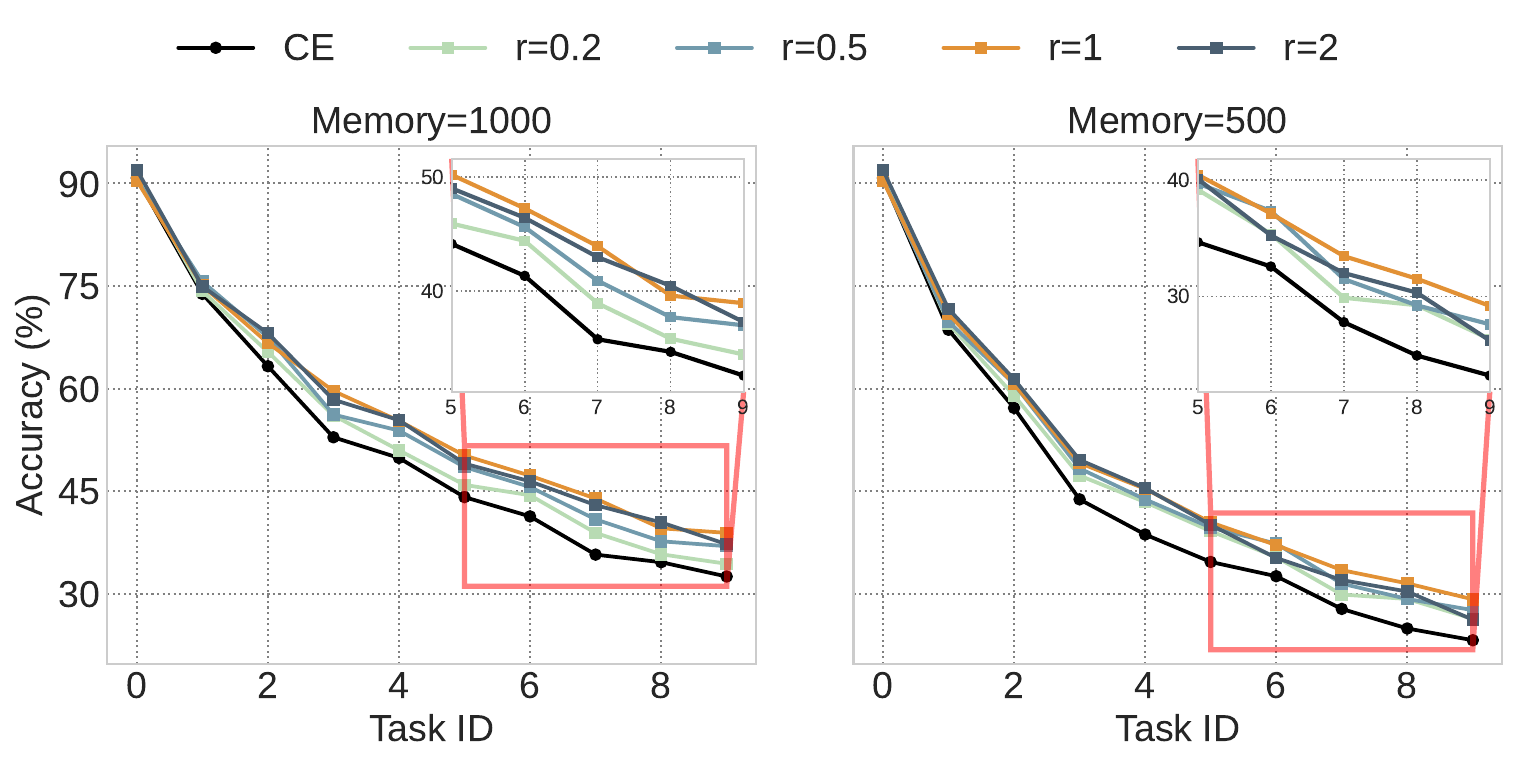}
    \caption{Forgetting curves with different replay buffer sizes}
    \label{fig:ablation1}
    \vspace{-2mm}
\end{figure}

Theoretically, the memory parameter $\lambda$ represents the estimated decay rate of supervision over time, which is closely related to both model architecture and dataset complexity. An excessively large or small $\lambda$ can lead to biased loss estimation. Tab.~\ref{tab:hyper_lambda_r} reports the grid search results over $\lambda$ and $r$, with a fixed replay buffer size of 2000 and an equal number of replay samples allocated to each old class. We observe that the optimal $\lambda$ in this experimental setting is 0.995. Moreover, TAL consistently outperforms the cross-entropy loss across a broad hyperparameter range, demonstrating its stability and robustness to hyperparameter choice.
Fig.~\ref{fig:ablation1} shows the forgetting curves under replay buffer sizes of 1000 and 500, with $\lambda = 0.999$ fixed for all subplots. TAL consistently raises the forgetting curves, indicating more stable retention over time. 

\begin{figure}[t!]
    \centering
    \includegraphics[width=1\linewidth, height=0.5\linewidth]{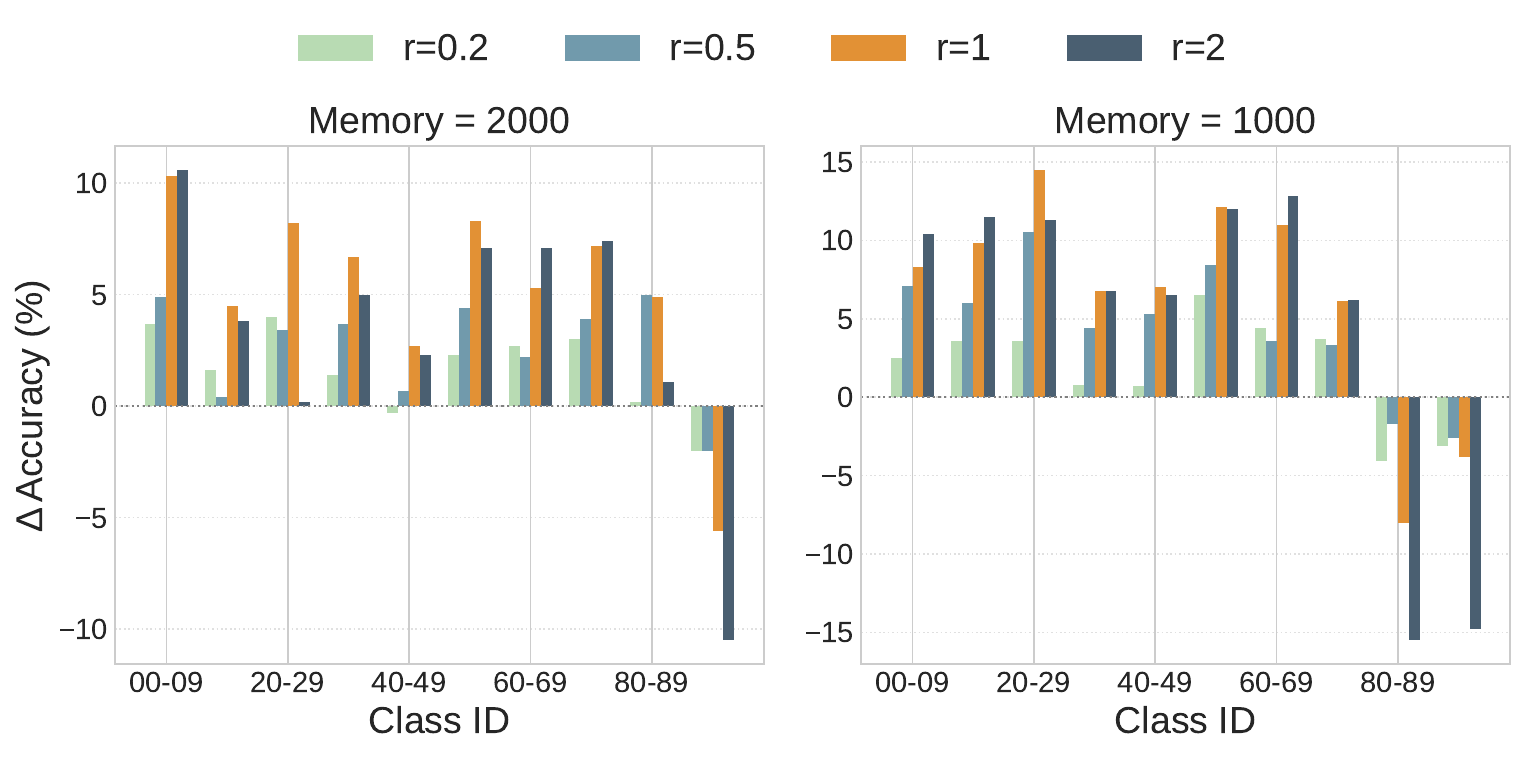}
    \caption{The impact of different $r$ values on the accuracy of each class, arranged in learning order at the end of the last task. Here, $\Delta$Accuracy is defined as the difference between the class accuracy of TAL and that of CE (\textit{i.e.}, $\Delta\text{Accuracy} = \text{Accuracy}_{\text{TAL}} - \text{Accuracy}_{\text{CE}}$). It can be observed that TAL does not simply apply uniform protection to all old classes. In the right figure, the more recent old classes (80–89) are even suppressed.}
    \label{fig:ablation2}
    \vspace{-3mm}
\end{figure}

The value of $r$ controls the relationship between the sensitivity to negative supervision and temporal positive supervision across classes. A larger $r$ makes the weight function $w(Q_k[N])$ sharper as $Q_k[N]$ increases, leading to stronger negative supervision on new classes, which enhances the protection of old classes but also limits the recall of new classes.  Fig.~\ref{fig:ablation2} illustrates the impact of different $r$ values on both new and old classes at the end of the last task. The experimental observations are consistent with our theoretical analysis: smaller $r$ values lead to weaker improvements for older classes and weaker suppression of newer classes, behaving more similarly to cross-entropy, whereas larger $r$ values amplify both the improvement of older classes and the suppression of newer classes.

\subsection{Further Analysis}

This section briefly discusses the applicability of TAL beyond traditional CIL settings and analyzes its \textbf{time efficiency}. 

According to the theoretical analysis in Sec.~\ref{method}, TAL degenerates to the standard cross-entropy loss under two idealized conditions: 
(1) the training data are perfectly class-balanced; 
and (2) the samples of each class arrive in a temporally uniform manner. However, in practice, real-world datasets are rarely class-balanced; moreover, even in standard supervised learning without CIL, the randomly shuffled training samples are often not strictly uniform over time. 
Based on these observations, we further evaluate the effectiveness of TAL under alternative settings, including:
\begin{itemize}
    \item CIL on long-tailed datasets \textbf{without exemplars}, using the \textbf{stronger pretrained ViT backbone}.
    \item comparison between TAL and CE in standard supervised learning \textbf{without CIL}.  
\end{itemize}
We also evaluate the time efficiency of TAL, and the results show that its additional computational overhead is minimal (about\textbf{ 0.8\%}).  

\textbf{Complete experimental results and detailed analyses are provided in the Appendix.}

\section{Conclusion}
\label{conclusion}

This work revisits the root causes of prediction bias in Class-Incremental Learning (CIL) and argues that, beyond the widely discussed intra-task class imbalance, the \emph{temporal imbalance} between positive and negative supervision plays a crucial role in driving models to suppress early classes and over-favor recently introduced classes. 
To address this, we propose \textbf{Temporal-Adjusted Loss (TAL)}, which performs dynamic temporal modeling for each class through a time-decay memory kernel. 
Theoretically, TAL degenerates to cross-entropy when sample arrivals are temporally uniform and the data are class-balanced, ensuring stability and compatibility. 
Empirically, TAL significantly mitigates forgetting and improves overall performance without modifying model architectures, yielding more persistent \emph{feature-level stability} rather than acting solely on the classifier head. 
Moreover, TAL is \textbf{plug-and-play} and can be seamlessly integrated into existing CIL frameworks.

\paragraph{Limitations.}
While TAL models temporal effects during training, it simplifies the decay of sample influence over time into a fixed exponential form. 
In reality, the memory parameter \(\lambda\) may vary across tasks or evolve dynamically over time. 
Additionally, the exponential decay approaches zero as time goes to infinity, which may not fully capture the persistence of learned representations. 
Future work will explore more flexible, non-parametric forms of temporal modeling to enhance both the interpretability and generality of TAL.

{
    \small
    \bibliographystyle{ieeenat_fullname}
    \bibliography{main}
}

\clearpage
\section{Appendix}

\subsection{Proof of Theorem~\ref{thm:backloading}}
\label{app:proof-backloading}

We provide a proof of Theorem~\ref{thm:backloading} under the notation used in Sec.~\ref{method}, where $N$ denotes the total number of steps, $a_k[n]\in\{+1,-1\}$ is the supervision polarity for class $k$ at step $n$, and $S_k[n]$ denotes the cumulative number of positive samples for class $k$ up to step $n$. Recall that
\begin{equation}
Q_k[N] = \sum_{n=0}^{N-1} f[N-1-n]\,a_k[n],
\label{eq:proof-start}
\end{equation}
where $f[\cdot]$ is a monotonically decreasing memory kernel with $f[n+1]\le f[n]$.

\paragraph{Step 1: Transforming the supervision polarity.}  
Let $b_k[n]=\frac{a_k[n]+1}{2}\in\{0,1\}$ so that $a_k[n]=2b_k[n]-1$. Substituting into \eqref{eq:proof-start} gives
\begin{equation}
Q_k[N] 
= 2\sum_{n=0}^{N-1} f[N-1-n]\,b_k[n] 
-\sum_{n=0}^{N-1} f[N-1-n].
\label{eq:proof-a2b}
\end{equation}

\paragraph{Step 2: Summation by parts.}  
Define the cumulative positive count
\[
S_k[n] := \sum_{j=0}^{n} b_k[j], \quad n=0,\dots,N-1,
\]
with $S_k[-1]:=0$.  
We apply summation by parts:
\begin{equation}
\begin{aligned}
&\sum_{n=0}^{N-1} f[N-1-n]\,b_k[n] = f[0]\,S_k[N-1] - \\
&\quad  \sum_{n=0}^{N-2} \big(f[N-2-n]-f[N-1-n]\big) S_k[n].
\end{aligned}
\label{eq:proof-parts}
\end{equation}

\paragraph{Step 3: Defining auxiliary statistics.}  
Let
\begin{equation}
\Delta_n := f[N-2-n]-f[N-1-n]\ge 0,\quad n=0,\dots,N-2,
\label{eq:proof-delta}
\end{equation}
and
\begin{equation}
\Phi_k(N) := f[0]\,S_k[N-1] - \sum_{n=0}^{N-2} \Delta_n\,S_k[n].
\label{eq:proof-phi}
\end{equation}
Then from \eqref{eq:proof-a2b} and \eqref{eq:proof-parts},
\begin{equation}
Q_k[N] = 2\,\Phi_k(N) - \sum_{n=0}^{N-1} f[N-1-n].
\label{eq:proof-q-phi}
\end{equation}

\paragraph{Step 4: Order equivalence.}  
The second term in \eqref{eq:proof-q-phi} is independent of $k$. Thus for any two classes $A$ and $B$,
\begin{equation}
Q_A[N] \;\le\; Q_B[N]
\quad\Longleftrightarrow\quad
\Phi_A(N) \;\le\; \Phi_B(N).
\label{eq:proof-order}
\end{equation}

\paragraph{Step 5: Equal positive counts.}  
Let $M_k := S_k[N-1]$ be the total number of positive samples for class $k$.  
If $M_A = M_B = M$, \eqref{eq:proof-phi} and \eqref{eq:proof-order} yield
\begin{equation}
Q_A[N] \;\le\; Q_B[N]
\quad\Longleftrightarrow\quad
\sum_{n=0}^{N-2} \Delta_n\, S_A[n] 
\;\ge\; \sum_{n=0}^{N-2} \Delta_n\, S_B[n].
\label{eq:proof-equal-M}
\end{equation}

\paragraph{Step 6: Back-loading condition.}  
Under the back-loading condition
\[
S_A[n] \ge S_B[n], \quad \forall n=0,\dots,N-2,
\]
and since $\Delta_n \ge 0$ by the monotonicity of $f[\cdot]$, it follows that
\[
\sum_{n=0}^{N-2} \Delta_n\, S_A[n] 
\;\ge\; \sum_{n=0}^{N-2} \Delta_n\, S_B[n],
\]
and therefore $Q_A[N] \le Q_B[N]$ by \eqref{eq:proof-equal-M}.

\paragraph{Step 7: Strict inequality.}  
If $f[\cdot]$ is strictly decreasing, then $\Delta_n>0$ for all $n$, and strict inequality holds whenever $\exists\,n$ such that $S_A[n] > S_B[n]$.

\paragraph{Conclusion.}  
This completes the proof of Theorem~\ref{thm:backloading}, showing that for two classes with the same total number of positive samples, the class with later (back-loaded) positives obtains a larger temporal supervision strength $Q_k[N]$.

\subsection{Invariant Range and Boundary Attainment of $Q$ and $w(\cdot)$}
\label{app:proof-invariance}

We show that the interval $[0,\,Q_{\max})$ is invariant for the recursion in Eq.~\ref{eq:markov-tal} and that $w(Q_k[N])\in[0,1)$ is preserved as well, together with the extremal sequences that attain the lower/upper boundary (in the appropriate sense).

\paragraph{Setup.}
Recall the recursion (Eq.~\ref{eq:markov-tal})
\begin{equation}
\begin{aligned}
Q_k[N+1]
= \lambda\Big(Q_k[N] + 
\begin{cases}
+1, & a_k[N]=+1,\\[2pt]
-\,w(Q_k[N]), & a_k[N]=-1,
\end{cases}
\Big),
\end{aligned}
\end{equation}
with \(0<\lambda<1\), \(Q_k[0]=0\), \(Q_{\max}=\frac{\lambda}{1-\lambda}\) and
\[
w(q)=\big(q/Q_{\max}\big)^{r}, \qquad r>0.
\]
Throughout we assume \(r\ge 1\) and \(Q_{\max}\ge 1\) (equivalently, \(\lambda\ge \tfrac12\)).

\newtheorem{proposition}{Proposition}
\begin{proposition}[Invariance of $[0,Q_{\max})$ and $[0,1)$]
\label{prop:invariance}

If \(0\le Q_k[N]<Q_{\max}\) and \(0\le w(Q_k[N])<1\), then 
\[
0\;\le\; Q_k[N+1]\;<\;Q_{\max},
\qquad
0\;\le\; w(Q_k[N+1])\;<\;1.
\]
\end{proposition}

\paragraph{Proof.}
\textit{Upper bound:} Assume \(Q_k[N]<Q_{\max}\).

(i) If \(a_k[N]=+1\),
\begin{equation}
\begin{aligned}
Q_k[N+1]
&= \lambda\big(Q_k[N]+1\big) \\
&\le \lambda(Q_{\max}+1) \\
&=\lambda\Big(\frac{\lambda}{1-\lambda}+1\Big) \\
&= Q_{\max},
\end{aligned}
\end{equation}
and the inequality is strict since \(\lambda<1\) and \(Q_k[N]<Q_{\max}\).

(ii) If \(a_k[N]=-1\), since \(0\le w(Q_k[N])<1\),
\begin{equation}
\begin{aligned}
Q_k[N+1]
&=\lambda\big(Q_k[N]-w(Q_k[N])\big) \\
&\le \lambda\,Q_k[N] \\
&< \lambda\,Q_{\max} \\
&< Q_{\max}.
\end{aligned}
\end{equation}

\textit{Lower bound:} Let \(q:=Q_k[N]\in[0,Q_{\max})\).  
With \(r\ge 1\) and \(Q_{\max}\ge 1\),
\begin{equation}
w(q)=\Big(\frac{q}{Q_{\max}}\Big)^{r}
\;\le\; \frac{q}{Q_{\max}}
\;\le\; q,
\end{equation}
so \(q - w(q)\ge 0\).

(i) If \(a_k[N]=+1\),
\[
Q_k[N+1] = \lambda(q+1) \ge 0.
\]

(ii) If \(a_k[N]=-1\),
\[
Q_k[N+1] = \lambda\big(q - w(q)\big) \ge 0.
\]

Thus, \(0\le Q_k[N+1]<Q_{\max}\).  
Finally, since \(Q_k[N+1]/Q_{\max}\in[0,1)\) and \(r>0\),
\begin{equation}
0\;\le\; w\big(Q_k[N+1]\big)
=\Big(\frac{Q_k[N+1]}{Q_{\max}}\Big)^{r}
<1.
\end{equation}
\hfill$\square$

\paragraph{Boundary attainment.}

\textit{Lower boundary (attained).}  
If \(a_k[n]\equiv -1\) for all \(n\ge 0\), then
\begin{equation}
\begin{aligned}
Q_k[1]&=\lambda(0-0)=0,\\
Q_k[2]&=\lambda(0-0)=0,\\
&\vdots
\end{aligned}
\end{equation}
so \(Q_k[N]\equiv 0\) for all \(N\), and \(w(Q_k[N])\equiv 0\).

\medskip
\textit{Upper boundary (approached, not attained).}  
If \(a_k[n]\equiv +1\) for all \(n\ge 0\),
\begin{equation}
Q_k[N+1]=\lambda\big(Q_k[N]+1\big), \qquad Q_k[0]=0.
\end{equation}
The closed-form solution is
\begin{equation}
\begin{aligned}
Q_k[N]
&=\lambda\sum_{m=0}^{N-1}\lambda^{m} \\
&=\frac{\lambda(1-\lambda^{N})}{1-\lambda} \\
&=Q_{\max}\big(1-\lambda^{N}\big).
\end{aligned}
\end{equation}
Hence \(Q_k[N]\ne Q_{\max}\) for finite \(N\), but
\[
Q_k[N]\;\uparrow\; Q_{\max}\qquad \text{as } N\to\infty.
\]
Correspondingly,
\begin{equation}
w\big(Q_k[N]\big)
=\Big(1-\lambda^{N}\Big)^{r}
\xrightarrow[N\to\infty]{} 1^{-}.
\end{equation}

\paragraph{Conclusion.}
Under the stated conditions, the recursion preserves
\[
Q_k[N]\in[0,Q_{\max}), \qquad
w(Q_k[N])\in[0,1),
\]
for all \(N\). The all-negative sequence attains the lower boundary exactly, while the all-positive sequence approaches the upper boundary asymptotically and never reaches it for \(0<\lambda<1\).

\subsection{Calibration Proof and Closed-Form \texorpdfstring{$\alpha$}{alpha} for \texorpdfstring{$r\in\{1,2\}$}{r=1,2}}
\label{app:alpha-proof}

Recall the steady-state calibration under temporally uniform and class-balanced data with prior $p=1/C$:
\[
\alpha \cdot w(Q^\ast) = 1, 
\qquad
w(q)=\Big(\frac{q}{Q_{\max}}\Big)^{r},
\]
and the steady-state recursion (taking expectation in Eq.~\ref{eq:markov-tal} and imposing $Q_k[N+1]=Q_k[N]=Q^\ast$):
\begin{equation}
Q^\ast = \lambda\Big(Q^\ast + p - (1-p)\,w(Q^\ast)\Big).
\label{eq:ss_Q_again}
\end{equation}
Let $x^\ast := Q^\ast/Q_{\max}\in(0,1)$ with $Q_{\max}=\lambda/(1-\lambda)$. Dividing \eqref{eq:ss_Q_again} by $Q_{\max}$ yields
\begin{equation}
(1-p)\,(x^\ast)^r + x^\ast - p = 0.
\label{eq:ss_poly_general}
\end{equation}
For $p=1/C$ this becomes
\begin{equation}
\Big(1-\frac{1}{C}\Big)(x^\ast)^r + x^\ast - \frac{1}{C} = 0,
\qquad
\alpha \;=\; \frac{1}{(x^\ast)^r}.
\label{eq:ss_poly_C_explicit}
\end{equation}

\paragraph{Existence and uniqueness of $x^\ast\in(0,1)$.}
Define $g(x)=(1-p)x^r + x - p$ on $[0,1]$. Then $g(0)=-p<0$, $g(1)=2(1-p)>0$ (for $C\ge 2$), and $g'(x)=(1-p)r x^{r-1}+1>0$. Hence $g$ is strictly increasing and crosses zero exactly once on $(0,1)$; therefore $x^\ast$ and $\alpha=1/(x^\ast)^r$ are unique.

\paragraph{Closed-form for $r=1$.}
When $r=1$, \eqref{eq:ss_poly_C_explicit} is linear:
\[
\Big(1-\frac{1}{C}\Big)x^\ast + x^\ast - \frac{1}{C}=0
\;\Longrightarrow\;
x^\ast=\frac{1/C}{\,2-1/C\,}=\frac{1}{\,2C-1\,}.
\]
Hence
\[
\alpha \;=\; \frac{1}{x^\ast} \;=\; 2C - 1.
\]

\paragraph{Closed-form for $r=2$.}
When $r=2$, \eqref{eq:ss_poly_C_explicit} is quadratic:
\[
\Big(1-\frac{1}{C}\Big)(x^\ast)^2 + x^\ast - \frac{1}{C}=0.
\]
The root in $(0,1)$ is
\[
x^\ast \;=\; \frac{-1 + \sqrt{\,1 + 4\frac{1}{C}\Big(1-\frac{1}{C}\Big)\,}}{\,2\Big(1-\frac{1}{C}\Big)\,}
\;=\; \frac{-C + \sqrt{\,C^2 + 4C - 4\,}}{\,2(C-1)\,}.
\]
Therefore
\[
\alpha \;=\; \frac{1}{(x^\ast)^2}
\;=\; \Bigg(\frac{C + \sqrt{\,C^2 + 4C - 4\,}}{2}\Bigg)^{\!2}.
\]

\paragraph{Other $r$.}
For general $r>4$, \eqref{eq:ss_poly_C_explicit} does not admit an elementary closed form in general. Since the left-hand side is strictly increasing on $[0,1]$ and changes sign between $0$ and $1$, one can solve for $x^\ast$ robustly by bracketing (e.g., bisection) or use Newton’s method
\[
x^{(t+1)}
= x^{(t)}
-\frac{\Big(1-\frac{1}{C}\Big)(x^{(t)})^{r} + x^{(t)} - \frac{1}{C}}
{\Big(1-\frac{1}{C}\Big) r (x^{(t)})^{r-1} + 1},
\]
initialized, e.g., at $x^{(0)}=\frac{1}{C}$. The unique solution $x^\ast\in(0,1)$ then yields $\alpha=1/(x^\ast)^r$.

\subsection{Pseudo Code of TAL}
\label{sec:pseudocode}

In this section, we summarize the batched implementation of the proposed \textbf{Temporal-Adjusted Loss (TAL)} in a clean and structured manner.  
Given a minibatch with logits \(Z \in \mathbb{R}^{N \times C}\) and labels \(y \in \{1,\dots,C\}^N\), TAL dynamically adjusts the supervision strength for negative samples based on the temporal supervision vector \(\mathbf{Q}\). Specifically, the loss reweights \emph{only} the negative logits using the multiplicative factor \(\alpha\,w(Q_k)\), where \(\alpha\) is determined by the number of classes \(C\) and the temporal adjustment parameter \(r\).  

\(\mathbf{Q}\) is initialized as a zero vector at the beginning of training and is updated online after every minibatch to track the relative temporal frequency of each class. The algorithm~\ref{alg:tal_tace_two_parts} below illustrates the forward loss computation and the update of \(\mathbf{Q}\) in each training step.

\begin{algorithm*}[h]
\caption{\textbf{TAL's Pseudo-code (Vectorized Form)}}
\label{alg:tal_tace_two_parts}
\begin{algorithmic}[1]
\Require Logits $Z \in \mathbb{R}^{N \times C}$, targets $y \in \{1,\dots,C\}^N$, 
         temporal supervision vector $\mathbf{Q}=(Q_1,\dots,Q_C)^\top \in \mathbb{R}^C$, 
         hyperparameters $\lambda \in (0,1)$, $r > 0$, $\alpha$ determined by $C$ and $r$
\Statex
\Statex \textbf{Initialization:}
\State $\mathbf{Q} \gets \mathbf{0}$ \Comment{Initialize $\mathbf{Q}$ as a zero vector at the beginning of training}
\State $Q_{\max} \gets \dfrac{\lambda}{1-\lambda}$

\Statex
\Statex \textbf{Part A: Loss computation}
\State $s_k \gets \Big(\dfrac{Q_k}{Q_{\max}}\Big)^{r}$ for $k=1,\dots,C$ \Comment{$s_k \in [0,1)$}
\State $\ell_k \gets \log\!\big(\alpha \cdot \max(s_k,\varepsilon)\big)$ for $k=1,\dots,C$ \Comment{stabilized log-weight}
\State $\widetilde{Z}_{i,k} \gets Z_{i,k} + \ell_k$ for all $i=1,\dots,N$ and $k=1,\dots,C$
\State $\widetilde{Z}_{i, y_i} \gets Z_{i, y_i}$ for all $i=1,\dots,N$ \Comment{vectorized replacement for true classes}
\State $\displaystyle 
\mathcal{L}_{\text{TAL}} \gets \frac{1}{N} 
\sum_{i=1}^N \left[
\log \left( \sum_{k=1}^C e^{\widetilde{Z}_{i,k}} \right) - Z_{i,y_i}
\right]$

\Statex
\Statex \textbf{Part B: Update of $\mathbf{Q}$}
\State $N_p(k) \gets \sum_{i=1}^N \mathbbm{1}\{\,y_i = k\,\}$ for $k = 1,\dots,C$
\State $N_n(k) \gets N - N_p(k)$ for $k=1,\dots,C$
\State \textit{Vector form:}\quad 
$\displaystyle \mathbf{Q} \gets \lambda \left( \mathbf{Q} + \frac{\mathbf{N}_p}{N} - \frac{\mathbf{N}_n}{N}\cdot \mathbf{s} \right)$
\State \textit{Component-wise (for clarity):}\quad 
$\displaystyle Q_k \gets \lambda \left( Q_k + \frac{N_p(k)}{N} - \frac{N_n(k)}{N}\cdot s_k \right),\quad k=1,\dots,C$
\State \Return $\mathcal{L}_{\text{TAL}}, \mathbf{Q}$
\end{algorithmic}
\end{algorithm*}

\subsection{Time Efficiency}
\label{sec:time_efficiency}

\subsubsection{Theoretical Complexity Analysis}
The proposed Temporal-Adjusted Loss (TAL) introduces two additional computations on top of the standard Cross-Entropy (CE) loss: (i) a temporal reweighting of negative logits via elementwise vector operations, and (ii) an online update of the temporal supervision vector $\mathbf{Q}$ after each minibatch.  
Let $N$ denote the batch size and $C$ the number of classes.

\paragraph{Cross-Entropy (CE).}
The standard CE loss involves computing the log-sum-exp across all $C$ classes for each of the $N$ samples:
\[
\mathcal{O}_{\text{CE}} = \mathcal{O}(NC),
\]
dominated by the log-sum-exp and gather operations.

\paragraph{TAL.} 
Compared to CE, TAL adds:
(1) a single vectorized elementwise operation to compute
\(\mathbf{s} = (Q/Q_{\max})^r\) and \(\ell = \log(\alpha \mathbf{s})\), and
(2) a vector update of $\mathbf{Q}$ using minibatch counts.
Both steps are $\mathcal{O}(C)$ operations, which are negligible compared to $\mathcal{O}(NC)$. \textbf{Thus, the total complexity is:}
\[
\mathcal{O}_{\text{TAL}} = \mathcal{O}(NC) + \mathcal{O}(C) = \mathcal{O}(NC),
\]
\textbf{identical in order to CE.} 

\subsubsection{Total Training Time Evaluation}
To assess the actual runtime impact of TAL in a realistic setting, we measure the \emph{total wall-clock training time} of several representative CIL methods on CIFAR-100, with and without TAL (i.e., using standard CE loss).  
This includes all forward and backward passes, model expansions, memory operations, and optimizer updates across the full training process.

We use the following setting:
\begin{itemize}
    \item \textbf{Dataset:} CIFAR-100
    \item \textbf{Number of classes per task:} 10
    \item \textbf{Exemplars per class:} 20
    \item \textbf{Batch size:} 128
    \item \textbf{Backbone:} ResNet-32
    \item \textbf{GPU:} NVIDIA A40, 46 GB
\end{itemize}

The results in Table~\ref{tab:time_ce_tal_cifar} show that integrating TAL into existing methods introduces only marginal runtime overhead. On average, the total training time increases by \(\mathbf{0.76\%}\).
This confirms that TAL’s lightweight computation has minimal impact at the full training scale.

\begin{table}[ht]
\centering
\caption{Comparison of \textbf{total training time} on CIFAR-100 between Cross-Entropy (CE) and TAL. The third column reports the relative change (\%) in total training time.}
\begin{tabular}{l|cc|c}
\toprule
\textbf{Method} & \textbf{CE (min)} & \textbf{TAL (min)} & \textbf{\% $\Delta$ (TAL vs CE)} \\
\midrule
ER      & 42.5  & 42.8  & +0.7\% \\
BiC     & 208.9 & 211.7 & +1.3\%  \\
iCaRL   & 44.2  & 45.2  & +2.3\%  \\
FOSTER  & 66.7  & 67.8  &  +1.6\% \\
MEMO    & 79.6  & 78.1  &  -1.9\%  \\
DER     & 81.3  & 82.1  & +1.0\%  \\
TagFex  & 125.1 & 125.1 &  0.0\%  \\
\midrule
\textbf{Average} & 92.6 & 93.3 &  +0.76\% \\
\bottomrule
\end{tabular}
\label{tab:time_ce_tal_cifar}
\end{table}

\noindent \textbf{Observation.}  
TAL adds less than \(1\%\) overhead on average over a complete training run, demonstrating that its temporal supervision mechanism does not slow down practical CIL pipelines. Meanwhile, Meanwhile, we also observe that BiC, which relies on balanced finetuning, significantly increases the total training time and computational cost. This highlights the advantage of TAL in correcting bias without requiring any additional post-processing.

\subsubsection{Loss-Only Runtime Evaluation}
While the previous experiment evaluates full training time, we further isolate the \emph{loss computation step} to examine TAL’s intrinsic overhead compared to CE.  
Specifically, we benchmark the per-batch time of CE and TAL loss computation under different batch sizes and numbers of classes, \emph{excluding forward propagation and any CIL-specific components} such as memory replay or model updates.  
This allows us to measure the pure cost of the loss function itself.

As shown in Table~\ref{tab:time_ce_tal}, TAL is slightly slower than CE due to the additional $\mathcal{O}(C)$ vector operations, but the gap remains constant and negligible across both batch size and class count. This empirically verifies the theoretical complexity analysis in Section~\ref{sec:time_efficiency}.

\begin{table}[ht]
\centering
\caption{\textbf{Per-batch loss computation time} comparison between Cross-Entropy (CE) and TAL under different batch sizes and numbers of classes. Units in $10^{-2}$\,ms.}
\resizebox{\linewidth}{!}{
\begin{tabular}{c|c|cccc}
\toprule
\textbf{Batch} & \textbf{Method} & \textbf{5 cls} & \textbf{20 cls} & \textbf{100 cls} & \textbf{500 cls} \\
\midrule
\multirow{2}{*}{32} 
& CE  & 10.89 & 10.78 & 10.77 & 10.81 \\
& TAL & 42.77 & 42.87 & 42.77 & 43.03 \\
\midrule
\multirow{2}{*}{64} 
& CE  & 10.83 & 10.86 & 10.76 & 10.81 \\
& TAL & 42.90 & 42.85 & 43.13 & 43.04 \\
\midrule
\multirow{2}{*}{128} 
& CE  & 10.72 & 10.74 & 10.75 & 10.79 \\
& TAL & 42.89 & 42.80 & 42.91 & 43.63 \\
\midrule
\multirow{2}{*}{256} 
& CE  & 10.80 & 10.81 & 10.82 & 11.08 \\
& TAL & 42.90 & 42.81 & 43.30 & 43.85 \\
\bottomrule
\end{tabular}}
\label{tab:time_ce_tal}
\end{table}

\noindent \textbf{Observation.}  
Even in this isolated setting, TAL remains efficient. The per-batch runtime difference does not scale with batch size or number of classes, confirming its constant and minimal overhead.

\subsection{TAL with Pretrained Model Based CIL Methods}
\label{sec:ptm_tal}

This section examines the performance of \textbf{TAL when applied to Pretrained Model Based CIL methods (PTM methods)}.  
It is important to note that \textbf{this is not a primary contribution of our work; instead, this experiment serves to demonstrate the effectiveness and applicability of TAL under different continual learning settings}.

In the main experiments of this paper, we assumed no pretrained backbone and trained models from scratch.  
In contrast, PTM methods rely on a \emph{pretrained vision transformer backbone}, where the feature extractor is pretrained and frozen throughout continual learning, eliminating feature drift between tasks.

Moreover, PTM methods are typically \emph{exemplar-free}. This implies that:
\begin{itemize}
    \item Each task is treated as an independent classification problem.
    \item Old classes are excluded from the loss computation during subsequent tasks.
    \item Each class receives supervision only in its original task and no further positive or negative supervision thereafter.
\end{itemize}
As a result, the \emph{positive-negative supervision imbalance} analyzed in Sec.~\ref{method} does not apply under the PTM setting.

However, as discussed in Sec.~\ref{conclusion}, TAL reduces to CE only if two conditions are met:
(1) the training samples of each class are temporally uniform, and
(2) the dataset is class-balanced.
While condition (1) is satisfied in PTM scenarios, condition (2) often is not—meaning that TAL remains effective for class-imbalanced datasets.

Tab.~\ref{tab:pretrained} reports results on three long-tailed datasets under 5-task and 10-task settings.  
We observe consistent improvements in both overall accuracy and tail class accuracy across multiple PTM-based baselines, demonstrating the general applicability of TAL even in pretrained settings without exemplars. $A_{\text{Mean}}$ denotes the average accuracy over all classes, while $A_{\text{Tail}}$ measures the accuracy on the tail classes, defined as the bottom 50\% of classes ranked by the number of training samples. All numbers are averaged on 3 different random seeds. Same as the Sec.~\ref{main_exp}, the TAL hyperparameters are fixed as $\lambda = 0.995$ and $r = 1$.

\begin{table*}[!ht]
\centering
\caption{Average accuracy ($A_{Mean}$) and tail class accuracy ($A_{Tail}$) under 5-task and 10-task settings across three long-tailed datasets. +TAL implies that cross entropy loss is replaced with TAL in the PTM-based methods.}
\scalebox{0.85}{
\begin{tabular}{l
  cc cc   cc cc   cc cc}
\toprule
\multirow{2}{*}{Method} 
  & \multicolumn{4}{c}{CUB-LT~\cite{wah2011caltech}}
  & \multicolumn{4}{c}{ImageNetR-LT~\cite{hendrycks2021many}}
  & \multicolumn{4}{c}{Food101-LT~\cite{he2025longtailedcontinuallearningvisual}} \\ 
\cmidrule(lr){2-5} \cmidrule(lr){6-9} \cmidrule(lr){10-13}
  & \multicolumn{2}{c}{5-task} 
  & \multicolumn{2}{c}{10-task} 
  & \multicolumn{2}{c}{5-task} 
  & \multicolumn{2}{c}{10-task} 
  & \multicolumn{2}{c}{5-task} 
  & \multicolumn{2}{c}{10-task} \\
\cmidrule(lr){2-3}\cmidrule(lr){4-5}
\cmidrule(lr){6-7}\cmidrule(lr){8-9}
\cmidrule(lr){10-11}\cmidrule(lr){12-13}
  & $A_{Mean}$ & $A_{Tail}$ & $A_{Mean}$ & $A_{Tail}$
  & $A_{Mean}$ & $A_{Tail}$ & $A_{Mean}$ & $A_{Tail}$
  & $A_{Mean}$ & $A_{Tail}$ & $A_{Mean}$ & $A_{Tail}$ \\ 
\midrule
L2P (CVPR'22)\cite{wang2022learning}             &  74.22   &  57.20   & 68.91  &  46.05  &  70.03 &  56.32    & 69.94   &  54.52      &  55.36   &  29.33   &   47.46  &  16.11  \\
+ TAL                     &  \textbf{75.91}   & \textbf{ 61.19}   &\textbf{ 70.71}  &  \textbf{49.80}  & \textbf{ 72.84 }& \underline{\textbf{ 63.50}}    &\textbf{ 71.63}   & \underline{\textbf{ 60.47}}      & \textbf{ 58.53 }  & \textbf{ 40.93}   & \textbf{ 50.84}   &  \textbf{23.84}  \\
\midrule
DualPrompt (ECCV'22)\cite{wang2022dualprompt}      &  73.85   &  51.30   & 69.87  &  50.75  &  65.40 &  48.28    & 65.14   &  46.52      &  52.24   &  25.55   &  49.84   &  22.56  \\
+ TAL                     &  \textbf{75.06}   & \textbf{ 54.65}   & \textbf{70.68}  &  \textbf{52.50}  & \textbf{ 66.45} &  \textbf{52.04}    &\textbf{ 66.00}   &  \textbf{50.06}      &  \textbf{54.28}   & \textbf{ 30.01}   &  \textbf{50.77}   &  \textbf{24.89}  \\
\midrule
CODAPrompt (CVPR'23)\cite{smith2023coda}      &  74.14   &  53.26   & 69.10  &  49.54  &  72.92 &  57.16    & 70.61   &  53.60      &  62.64   &  34.71   &  56.11   &  28.33  \\
+ TAL                     & \textbf{ 75.51}   &  \textbf{56.04}   & \textbf{70.45 } & \textbf{ 51.68}  & \underline{\textbf{ 75.34}} & \textbf{ 62.08}    & \underline{\textbf{72.86}}   &  \textbf{58.00 }     & \underline{\textbf{ 65.19 }}  &  \textbf{39.44}   &  \textbf{58.12}   & \textbf{ 34.10}  \\
\midrule
LAE (ICCV'23) \cite{gao2023unified}            &  74.54   &  57.90   & 72.03  &  50.20  &  69.13 &  53.52    & 69.54   &  56.00      &  62.14   &  37.96   &  56.43   &  29.91  \\
+ TAL                     &  \underline{\textbf{76.97}}   & \underline{\textbf{ 61.40}}   &\underline{\textbf{ 74.83}}  &  \underline{\textbf{57.10}}  & \textbf{ 70.54} &  \textbf{58.48 }   & \textbf{71.00 }  &  \textbf{58.64 }     &  \textbf{64.34 }  & \underline{\textbf{ 42.35 }}  & \underline{\textbf{ 59.38}}   & \underline{\textbf{ 35.89 }} \\
\midrule
MOS (AAAI'25)\cite{sun2025mos}             &  73.95   &  51.40   & 71.52  &  48.93  &  69.15 &  53.80    & 68.73   &  55.44      &  60.63   &  29.30   &  55.46   &  28.08  \\
+ TAL                     &  \textbf{75.30}   & \textbf{} \textbf{54.86}   &\textbf{ 72.14}  &  \textbf{50.95}  & \textbf{ 71.47} &  \textbf{59.29}    &\textbf{ 70.31 }  &  \textbf{58.77}      & \textbf{ 63.24}   & \textbf{ 33.27}   &  \textbf{57.92}   &  \textbf{33.73}  \\
\bottomrule
\end{tabular}
}
\label{tab:pretrained}
\end{table*}

\begin{figure}[h]
    \centering
    \includegraphics[width=0.45\textwidth, height=0.6\textheight]{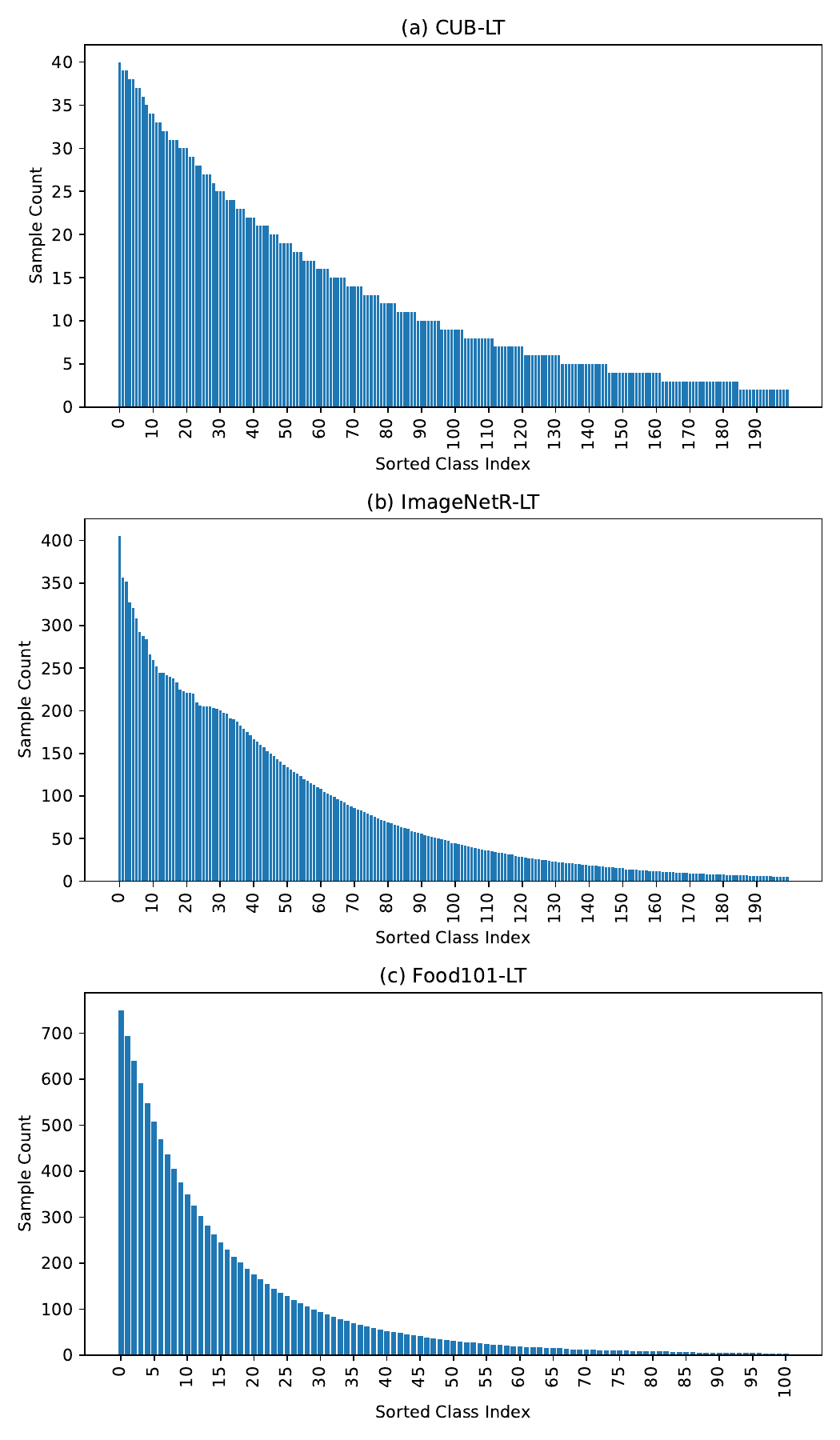}
    \caption{Sample-Class Distribution of Long-Tailed Datasets}
    \label{fig:longtailed_dataset}
    \vspace{-2mm}
\end{figure}

\paragraph{Implementation Details.}
For experiments with PTM methods, as shown in~\ref{fig:longtailed_dataset}, we adopt three long-tailed datasets: \textbf{CUB-LT}~\cite{wah2011caltech}, \textbf{ImageNetR-LT}~\cite{hendrycks2021many}, and \textbf{Food101-LT}~\cite{he2025longtailedcontinuallearningvisual}. 
Each long-tailed training set is constructed by resampling the original dataset following a Pareto distribution with shape parameter $\alpha = 6$, as in~\cite{liu2019large}. 
The imbalance factors $\rho = n_{\max}/n_{\min}$ are set to 20 for CUB-LT, 81 for ImageNetR-LT, and 150 for Food101-LT. 
The test sets are class-balanced, containing 20 samples per class for CUB-LT and 25 samples per class for ImageNetR-LT and Food101-LT.

All PTM experiments use a \textbf{frozen pretrained ViT backbone}, following standard pretrained model-based CIL settings. 
No exemplars are used, and each task introduces a disjoint set of classes without overlap. 
We adopt the default hyperparameter configurations provided in the \textbf{LAMDA-PILOT}~\cite{sun2025pilot} framework for all methods, including learning rate, optimizer settings, and training schedule. 
The number of tuning epochs for each task is fixed at 20 across all PTM methods to ensure a fair comparison.

\noindent\paragraph{Observations} Although PTM settings do not involve positive-negative supervision imbalance for the same class across different tasks, tail classes within each task still suffer from excessively strong negative supervision. TAL addresses this issue without relying on any prior class knowledge or post-processing, and consistently improves both $A_{Mean}$ and $A_{Tail}$ on long-tailed datasets. This demonstrates that TAL is effective beyond non-PTM scenarios and can be seamlessly integrated into pretrained pipelines without architectural or training modifications. \textbf{Again, this section is intended only to showcase the future potential of TAL in different continual learning settings, rather than being claimed as a main contribution of this work.}

\subsection{TAL in Standard Supervised Learning}
\label{sec:tal_supervised}

As discussed in Sec.~\ref{conclusion}, TAL reduces to CE only if two conditions are met:
(1) the training samples of each class are temporally uniform, and
(2) the dataset is class-balanced.
This implies that if both conditions hold---as in standard supervised learning settings where samples are randomly shuffled and the class distribution is balanced---TAL should theoretically behave similarly to CE.

However, we consistently observe in practice that even on the first task, where no forgetting can occur, TAL often yields slightly higher test accuracy than CE when appropriately tuned. 
This phenomenon caught our attention, as the first task represents a purely supervised learning scenario without any temporal imbalance.
To verify this effect, we conducted experiments on CIFAR-100 and ImageNet, where we randomly selected subsets of classes and trained models using ResNet-32 and ResNet-18 backbones from scratch, respectively. 
We used identical optimizers and training schedules for both CE and TAL to ensure a fair comparison. 

Tab.~\ref{tab:ce_tal_acc} summarizes the results under different batch sizes and class numbers. 
Across all configurations, TAL consistently outperforms CE by a small but stable margin. 
This suggests that TAL not only maintains theoretical equivalence to CE under balanced conditions but may also provide a mild regularization effect through its temporal weighting mechanism, leading to slightly improved generalization.
\begin{table}[ht]
\centering
\caption{Average accuracy (\%) comparison between Cross-Entropy (CE) and TAL under different batch sizes and class numbers on CIFAR and ImageNet. \textbf{Classes are randomly chosen with 10 different random seeds}.}
\resizebox{\columnwidth}{!}{
\begin{tabular}{c|c|ccc}
\toprule
\textbf{Dataset} & \textbf{Method} & \textbf{Batch=64} & \textbf{Batch=128} & \textbf{Batch=256} \\
\midrule
\multicolumn{5}{c}{\textbf{5 classes}} \\
\midrule
\multirow{2}{*}{CIFAR} 
& CE  & 88.56 & 88.68 & 85.88 \\
& TAL & \textbf{89.36} & \textbf{89.24} & \textbf{89.36} \\
\midrule
\multirow{2}{*}{ImageNet} 
& CE  & 96.83 & 94.84 & 96.43 \\
& TAL & \textbf{97.62} & \textbf{95.24} & \textbf{97.22} \\
\midrule
\multicolumn{5}{c}{\textbf{10 classes}} \\
\midrule
\multirow{2}{*}{CIFAR} 
& CE  & 85.64 & 84.56 & 84.28 \\
& TAL & \textbf{87.42} & \textbf{85.76} & \textbf{84.88} \\
\midrule
\multirow{2}{*}{ImageNet} 
& CE  & 89.00 & 86.60 & 88.24 \\
& TAL & \textbf{89.80} & \textbf{88.40} & \textbf{88.68} \\
\bottomrule
\end{tabular}}
\label{tab:ce_tal_acc}
\end{table}
We believe this observation opens up interesting directions for applying TAL beyond continual learning. 
One possible explanation for this phenomenon is the existence of a subtle inter-batch forgetting effect. 
Specifically, since each epoch in standard supervised learning involves random shuffling of the training data, some classes may have a larger proportion of their samples appearing in earlier batches of an epoch. 
As a result, by the end of the epoch, these classes may have received relatively stronger negative supervision, leading to a mild temporal imbalance effect that resembles, to a lesser degree, the temporal imbalance observed in continual learning scenarios.

This finding suggests that TAL’s temporal adjustment mechanism can have beneficial effects even in conventional supervised learning, by mitigating these subtle temporal biases during optimization. 
It is important to note, however, that this experiment is intended to \textbf{demonstrate the broader applicability and potential of TAL} in more general training settings and \textbf{is not claimed as a main contribution of this work}.

\end{document}